\documentclass[final]{IEEEtran}

\usepackage{amsmath,amssymb} 
\usepackage{color}
\usepackage{epsfig}
\usepackage{comment}
\usepackage[ruled,linesnumbered]{algorithm2e}
\usepackage{enumitem}
\usepackage[export]{adjustbox}
\usepackage{tabularx}
\usepackage[table]{xcolor}
\usepackage{colortbl}
\usepackage{soul}
\usepackage{multirow}
\usepackage{booktabs}
\usepackage{xr}
\usepackage{pifont}
\usepackage{hyperref}

\usepackage{sidecap}

\usepackage{ragged2e}

\usepackage{dsfont}

\usepackage{cite}

\ifCLASSINFOpdf
\else
\fi
\usepackage{amsmath}
\usepackage{array}

\ifCLASSOPTIONcompsoc
 \usepackage[caption=false,font=normalsize,labelfont=sf,textfont=sf]{subfig}
\else
 \usepackage[caption=false,font=footnotesize]{subfig}
\fi
\usepackage{url}

\begin{document}
%
\title{Multi-Target Multi-Camera Tracking of Vehicles using Metadata-Aided Re-ID and Trajectory-Based Camera Link Model}
%
%
%

\author{
Hung-Min~Hsu,~\IEEEmembership{Member, IEEE},
Jiarui Cai,~\IEEEmembership{Student Member, IEEE},
Yizhou Wang,~\IEEEmembership{Student Member, IEEE},
Jenq-Neng Hwang,~\IEEEmembership{Fellow, IEEE},
Kwang-Ju~Kim
\thanks{Hung-Min Hsu, Jiarui Cai, Yizhou Wang, and Jenq-Neng Hwang are with the Department
of Electrical and Computer Engineering, University of Washington, Seattle, WA 98195, USA. E-mail:\texttt{\{hmhsu, jrcai, ywang26, hwang\}@uw.edu}.}
\thanks{Kwang-Ju Kim is with Electronics and Telecommunications Research Institute (ETRI) Daegu-Gyeongbuk Research Center, 42994, South Korea. Email: \texttt{kwangju@etri.re.kr}.}
}

%
%

\markboth{IEEE Transactions on Image Processing}%
{Hsu \MakeLowercase{\textit{et al.}}: Multi-Target Multi-Camera Tracking of Vehicles using Metadata-Aided Re-ID and Trajectory-Based Camera Link Model}
%





\maketitle

\begin{abstract}
In this paper, we propose a novel framework for multi-target multi-camera tracking (MTMCT) of vehicles based on metadata-aided re-identification (MA-ReID) and the trajectory-based camera link model (TCLM). Given a video sequence and the corresponding frame-by-frame vehicle detections, we first address the isolated tracklets issue from single camera tracking (SCT) by the proposed traffic-aware single-camera tracking (TSCT). Then, after automatically constructing the TCLM, we solve MTMCT by the MA-ReID. The 
TCLM is generated from camera topological configuration to obtain the spatial and temporal information to improve the performance of MTMCT by reducing the candidate search of ReID. We also use the temporal attention model to create more discriminative embeddings of trajectories from each camera to achieve robust distance measures for vehicle ReID. Moreover, we train a metadata classifier for MTMCT to obtain the metadata feature, which is concatenated with the temporal attention based embeddings. Finally, the TCLM and hierarchical clustering are jointly applied for global ID assignment. The proposed method is evaluated on the CityFlow dataset, achieving IDF1 76.77\%, which outperforms the state-of-the-art MTMCT methods.
\end{abstract}


%


\section{Introduction}\label{sec:intro}

\begin{figure}[t]
\begin{center}
 \includegraphics[width=1.0\linewidth]{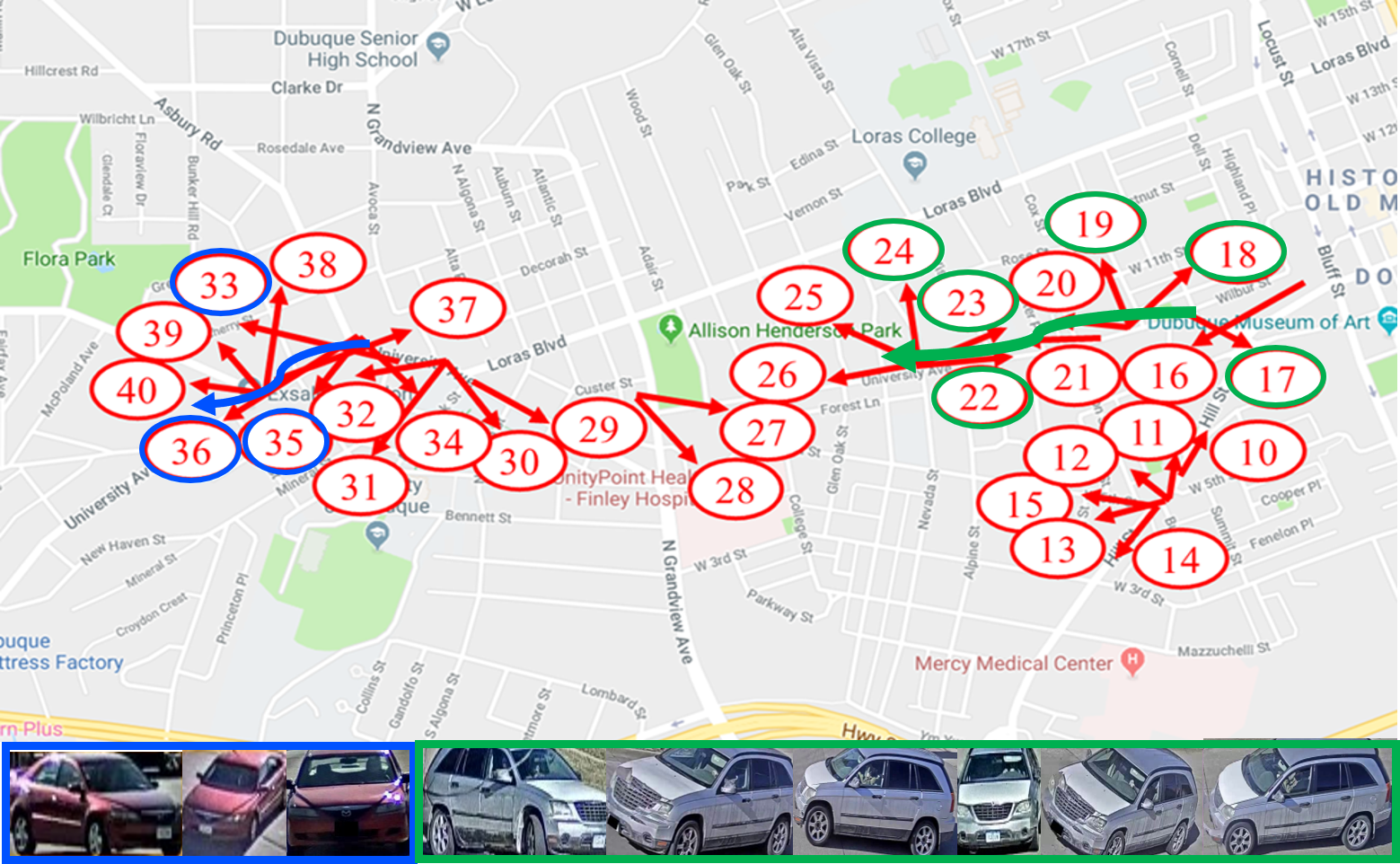}
\end{center}
 \caption{Illustration for MTMCT of vehicles. Given any vehicle in videos recorded by several time-synchronized cameras with/without overlapping field of views (FoVs), the MTMCT task is aimed to track the same vehicle in all the cameras.}
\label{fig:TNT_track1}
\end{figure}

\IEEEPARstart{D}{ue} to the exponential growth of intelligent transportation systems, multi-target multi-camera tracking is becoming one of the important tasks. The purpose of MTMCT is to identify and locate targets in a multi-camera system. For instance, Fig.~\ref{fig:TNT_track1} shows there are two vehicles tracked by an MTMCT system. However, there are some fundamental problems in detecting and tracking need to be solved to achieve this goal. Basically, most of the MTMCT systems are composed of two modules, i.e., single camera tracking (SCT) and inter-camera tracking (ICT). SCT aims to track the vehicle trajectories within a single camera. In terms of ICT, it is to re-identify the vehicle trajectories across multiple cameras by vehicle re-identification (ReID) \cite{tang2019pamtri}. 

\begin{figure*}
\centering
\includegraphics[width=0.8\textwidth]{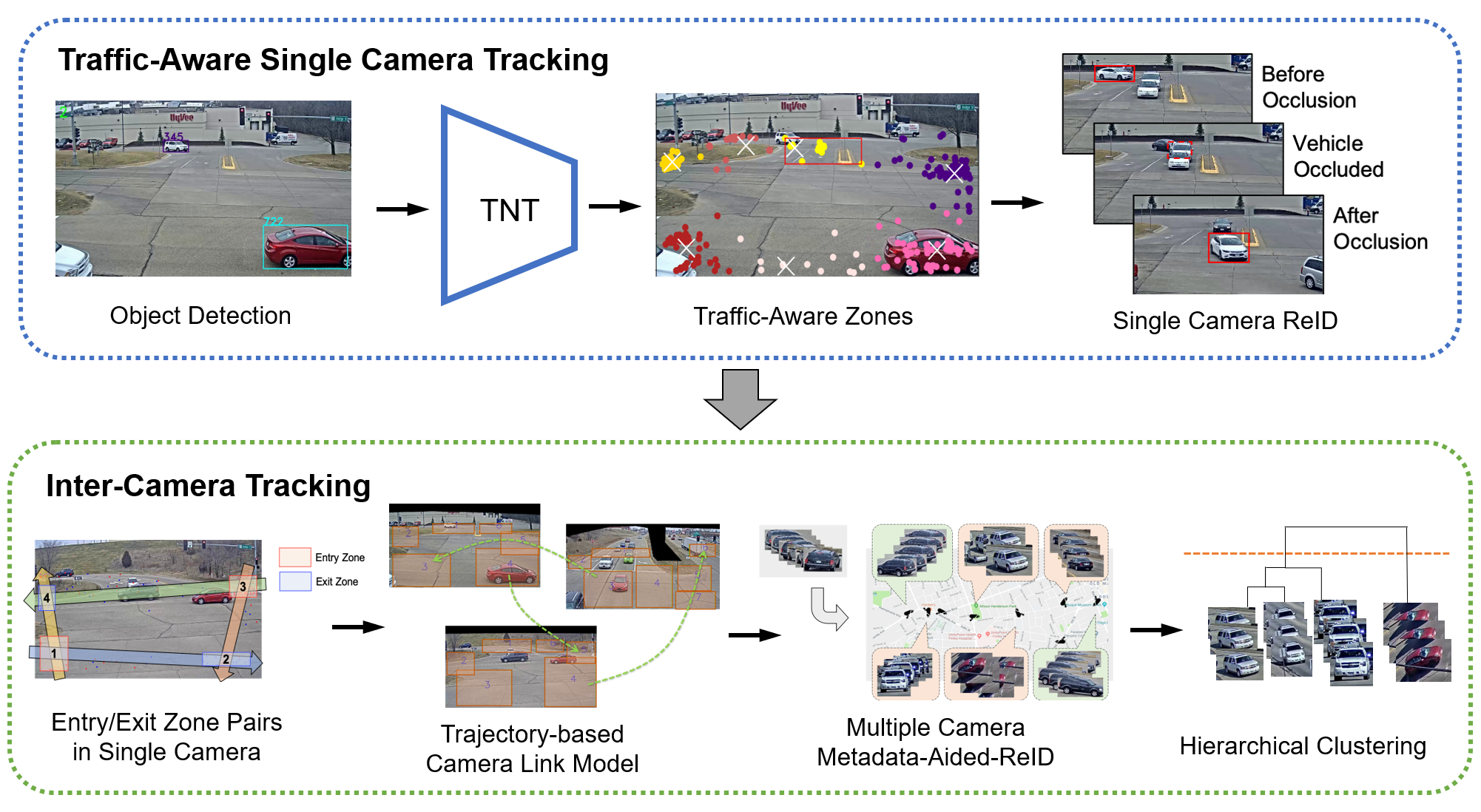}
\caption{The illustration of the proposed MTMCT framework. First, the TSCT is adopted as the single camera tracker to obtain SCT results so that entry, exit, and traffic-aware zones can be generated. Then, trajectory-based camera link model (TCLM) with transition time are automatically generated based on entry/exit zones and training data. Finally, metadata-aided ReID is applied to the solution space with the camera link constraint, and the hierarchical clustering is used for generating the final MTMCT results.}
\label{fig:framework}
\end{figure*}

In fact, MTMCT is a very complicated task since SCT and ReID are both challenging research topics. In SCT, the tracking-by-detection scheme has been proven effective in many works \cite{geiger20143d,zhang2013understanding}. Tracking-by-detection is composed of two steps: (i) object detection in each frame, (ii) trajectory generation by associating the corresponding detections across time. However, it is difficult to track vehicles within a single camera due to unreliable vehicle detection and heavy occlusion. For ReID, the different orientations of the same vehicle, the same car model from different vehicles, low resolution of video, and varying lighting conditions are all intractable problems. The poor performance of SCT and ReID can cause dramatic ID switching. If the trajectories are lost in SCT, ID switching will happen in MTMCT. On the other hand, the ReID model also needs the capability to identify the appearance of the same target in varied illuminations and viewing angles in different cameras to avoid ID switching. To alleviate the mentioned issues, a more robust appearance feature is critically needed for SCT and ReID. To connect the vehicular tracks cross cameras, appearance feature based vehicle ReID is one of the most effective approaches. For vehicle ReID, some works focus on generating discriminative features by deep convolutional neural networks (CNNs). In most of the methods, the trained ReID model is used to extract effective embedding features, which can be used to estimate the similarity based on Euclidean distance between trajectories in the testing stage. To the best of our knowledge, there is no MTMCT system using the metadata information, such as brand, type and color of vehicles for ICT. Nonetheless, many vehicles are of the same brand, type and color but these vehicles are not the same identity. Thus, we assert that not only metadata information but also spatial and temporal information are all critical information for MTMCT.

In this paper, we propose an innovative framework for MTMCT of vehicles to solve all the mentioned issues. The flowchart of our proposed MTMCT system is shown in Fig.~\ref{fig:framework}. First, we use a TrackletNet Tracker (TNT) \cite{wang2018exploit} in the SCT, which has a superior performance in many intelligent transportation system applications \cite{wang2019anomaly,wang2019monocular}. Then based on the appearance feature similarity and bounding box intersection-over-union (IOU) between consecutive frames, the detection results are associated into tracklets. For the neighboring disjoint tracklets, we estimate their similarity by a Siamese TrackletNet based on both appearance and temporal information. A graph model is then built with tracklets being treated as vertices and similarities between two tracklets as measured by the TrackletNet being treated as edge weights. Then the graph is partitioned into small groups, where each group can represent a unique vehicle ID and moving trajectory in each camera. To enhance the performance of SCT, traffic-aware single camera tracking (TSCT) is further proposed to solve the long-term occlusion due to the traffic scenario. Then, we combine the temporal attention weighted clip-level deep appearance feature and metadata of the vehicle as the auxiliary information to establish a more representation feature for ReID across different cameras \cite{huang2019multi}. Basically, there are three metadata information: car type, brand and color. Finally, we propose the trajectory-based camera link model (TCLM) to obtain the spatial and temporal information to reduce the ambiguity of the different identities with the same car model. By exploiting TCLM, the dependency of well-embedded appearance can be mitigated based on these spatial and temporal constraints. To summarize, we claim the following contributions,

\begin{itemize}
\item The traffic-aware single camera tracking (TSCT) is proposed to achieve the best performance in the MTMCT task.
\item Combining clip-level appearance feature and meta information to generate the feature of each trajectory for ReID across cameras.
\item Trajectory-based camera link model (TCLM) are constructed to exploit the spatial and temporal information to enhance the performance of MTMCT.
\end{itemize}

The rest of this paper is organized as follows. Section \ref{sec:related_works} reviews related works. Then, we elaborate the proposed MTMCT system in Section \ref{sec:method}. In Section \ref{sec:results}, we evaluate the proposed method on the CityFlow dataset \cite{tang2019cityflow,naphade20192019,naphade20204th} and compare it with the state-of-the-art methods. Finally, the paper is concluded in Section \ref{sec:conclusion}.

\section{Related Works}\label{sec:related_works}
A large amount of literature on person ReID and MTMCT have attracted growing attention in the past few years. In addition, some works tackle vehicle ReID due to smart-city-related applications. In this section, we discuss the most relevant research works to the MTMCT tasks by the following two parts: overlapping field of view (FOVs) and non-overlapping FOVs.

\subsection{Overlapping FOVs.} 
There are many research studies to solve the MTMCT tasks with overlapping FOVs between cameras. Fleuret et al. \cite{fleuret2007multicamera} use probabilistic occupancy map (POM) with color and motion attributes for MTMCT. Berclaz et al. \cite{berclaz2011multiple} formulate the MTMCT task as an integer programming problem and deal with the problem by the $k$-shortest paths (KSP) algorithm. Moreover, some research works use graph modeling algorithms to solve the MTMCT problem. For example, Hofmann et al. \cite{hofmann2013hypergraphs} and Shitrit et al. \cite{shitrit2013multi} use a constrained min-cost flow graph approach to associate the detections frame-by-frame. Leal et al. \cite{leal2012branch} formulate the MTMCT problem as a multi-commodity network flow problem and use the branch-and-price (B\&P) algorithm to link detections into trajectories.
Recently, the two-step approach for the MTMCT problem is becoming more and more popular. The main idea of the two-step approach is firstly to track all the targets within a single camera, then match the generated trajectories from every single camera to all the other cameras. Therefore, the first step is SCT, which is intensively studied in computer vision and pattern recognition communities \cite{yang2014robust,zhao2010human,ma2015hierarchical,lan2017learning,cai2020ia,zhang2020lifts}. Thanks to the great advances of convolution neural networks (CNNs), object detection techniques have been shown to achieve impressive performance in recent years, tracking-by-detection scheme \cite{yu2016poi,jiang2015online,zhu2018online,tang2017multiple,liu2019model} has thus become the mainstream approach for multiple object tracking. After SCT, the next step is to match local trajectories on different cameras. Hu et al. \cite{hu2006principal} and Eshel et al. \cite{eshel2008homography} associate the trajectories across cameras via an epipolar geometry constraint, which predicts the bounding boxes of those targets into a reference plane on the next frame and determines those targets with intersections in the reference plane as the same identity. Bredereck et al. \cite{bredereck2012data} propose a Greedy Matching Association (GMA) method to iteratively associate single-camera trajectories to generate the cross-camera trajectories. Xu et al. \cite{xu2016multi} propose a Hierarchical Composition of Tracklet (HCT) framework to match local tracklets to the other cameras by using the appearance feature and the ground plane locations. Xu et al. \cite{xu2017cross} further propose the Spatio-Temporal Parsing (STP) structure, which prunes matching candidates of trajectories by using semantic attributes of targets, to solve the tracklet matching problem.

\subsection{Non-overlapping FOVs.} 
The other category of research works that aim at the MTMCT task with non-overlapping FOVs. Research studies in this category attempt to match single-camera trajectories across different non-overlapping FOV cameras by exploiting different information such as appearance feature \cite{tesfaye2017multi,cai2014exploring}, motion pattern \cite{hofmann2013hypergraphs}, and camera topological configuration\cite{lee2017online}. 
For appearance cues, Cai et al. \cite{cai2014exploring} use a Relative Appearance Context (RAC) to differentiate adjacent targets. Chen et al. \cite{chen2014novel} exploit the Piecewise Major Color Spectrum Histogram Representation (PMCSHR) to estimate the similarity of targets in different views by generating a major color histogram for each target. Zhang et al. \cite{zhang2017multi} use Convolutional Neural Networks (CNNs) to generate the feature representation for each target and propose a Feature Re-Ranking mechanism (FRR) to find correspondences among tracklets. 
Ristani and Tomasi \cite{ristani2018features} consider not only the CNN based appearance feature but also motion pattern. Moreover, they formulate the MTMCT task as a binary integer program problem and propose deep feature correlation clustering (DeepCC) approach to match the trajectories of a single camera to all the other cameras. Chen et al. \cite{chen2016equalized} propose an Equalized Graph Model (EGM) to find the solution of trajectory assignment by defining a trajectory in each camera as a node in the graph and edges are the connections of trajectories. The final cross camera trajectory of each target is to find the min-cost flow from the source node to the sink one. Chen et al. \cite{chen2016integrating} propose social grouping in MTMCT, which uses Conditional Random Field (CRF) to match single-camera trajectories across different cameras by minimizing the unary and pairwise energy costs. Besides, Ye et al. \cite{ye2019dynamic} design a dynamic graph matching (DGM) framework for video-based person ReID, which utilizes a co-matching strategy to reduce the false matching in MTMCT.

Some research works consider the camera topology in MTMCT. For example, \cite{cheng2017part,nie2014single} attempt to match local tracklets between every two neighboring cameras. Lee et al. \cite{lee2017online,lee2017inter} present a fully unsupervised online learning approach, which efficiently integrates discriminative visual features by the proposed Two-Way Gaussian Mixture Model Fitting (2WGMMF) and context feature, to systematically build camera link model so as to match single-camera trajectories to the neighboring cameras. On the other hand, Wang et al. \cite{wang2020exploiting} propose consistent cross-view matching (CCM) framework to use the global camera network constraints for video-based person ReID, which also proves the effectiveness of using the camera topology for MTMCT.

\section{Proposed MTMCT Framework}\label{sec:method}

\begin{table}
\caption{The notation definition.}
\label{table}
\centering
\setlength{\tabcolsep}{3pt}
\begin{tabular}{|p{25pt}|p{180pt}|}
\hline
Symbol& 
Description\\
\hline






$c^{t}_{i}$&
A candidate centroid $i$ at iteration $t$.\\

$N(c^{t}_{i})$& 
The neighborhood of samples within a given distance around $c_{i}$.\\



$N_{e,k}$& 
Number of entry points in each zone.\\

$N_{x,k}$&
Number of exit points in each zone.\\

$D_e$&
Entry zone density.\\

$D_x$&
Exit zone density.\\

$D_{ta}$&
Traffic-aware zone density.\\











${\mathcal{M}(\xi_j)}$&
Metadata feature of a trajectory $\xi_j$.\\




$C^{s}$&
zone pair set in the source camera.\\

$C^{d}$&
zone pair set in the destination camera.\\

$P^{s}_{i}$&
$i$th zone pair in $C^{s}$.\\

$P^{d}_{j}$&
$j$th zone pair in $C^{d}$.\\

$z_{s}$&
Transition zone of $P^{s}_{i}$.\\

$z_{d}$&
Transition zone of $P^{d}_{j}$.\\

$\alpha_z$&
Overlapping ratio of the vehicle to zone $z$.\\


\hline

\end{tabular}
\label{tab:notation}
\end{table}

\begin{algorithm}[t]
\SetAlgoLined
\SetKwInOut{Input}{Input}
\SetKwInOut{Output}{Output}
\SetKw{KwFrom}{from}
\SetKw{KwAnd}{and}
\Input{Detections set $\mathcal{D}$ from all $V$ cameras.}
\Output{Global ID for all trajectories within all $V$ cameras.}

\For{camera $i$ \KwTo $V$}{
$\Xi^i \leftarrow$ TNT$(\mathcal{D}_i)$
\tcp{generate trajectories set $\Xi^i = \{\xi^i_n\}$ of camera $i$}
$Z^i \leftarrow$ MeanShift$(\Xi^i)$
\tcp{generate zones $Z^i$ from trajectories $\Xi^i$}
Calculate the traffic-aware zone density $D_{ta}$, exit density $D_x$ and entry density $D_e$ to classify $Z^i$ into the traffic-aware zone, exit zone and entry zone\;
Merge the isolated trajectories in the traffic-aware zones\;
\For{trajectory $\xi^i_j$ \KwFrom $\Xi^i$}{
$\mathcal{A}(\xi_j^i) \leftarrow$ TA-ReID$(\xi_j^i)$
\tcp{extract appearance embedding using the Temporal Attention ReID model}
$\mathcal{M}(\xi_j^i) \leftarrow$ MetadataClassifier$(\xi_j^i)$
\tcp{obtain metadata features}
$\mathbf{f}(\xi_j^i)=\mathcal{A}(\xi_j^i)\oplus\mathcal{M}(\xi_j^i)$\;
}
}
Trajectories set for all cameras $\mathcal{T} \leftarrow \left\{ \Xi^{1}, \Xi^{2}, \cdots, \Xi^{V} \right\}$\;
Global IDs $\leftarrow$ HierarchicalClustering$(\mathbf{f}(\mathcal{T}))$\;
\caption {The proposed MTMCT}
\label{code:whole_framework}
\end{algorithm}

There are four steps in the proposed MTMCT framework: (1) Apply traffic-aware single camera tracking (TSCT) to generate SCT results. (2) Train a ReID model and metadata classifier to extract the appearance feature and metadata feature of each trajectory. (3) Establish the TCLM (i.e., the spatial and temporal constraint) for ICT. (4) Use the feature of each trajectory and TCLM to generate the ICT results. We show the notation table in Table~\ref{tab:notation} and the whole framework in Alg.~\ref{code:whole_framework}.

\subsection{Traffic-Aware Single Camera Tracking}

\begin{figure}[t]
\centering
\includegraphics[width=\linewidth]{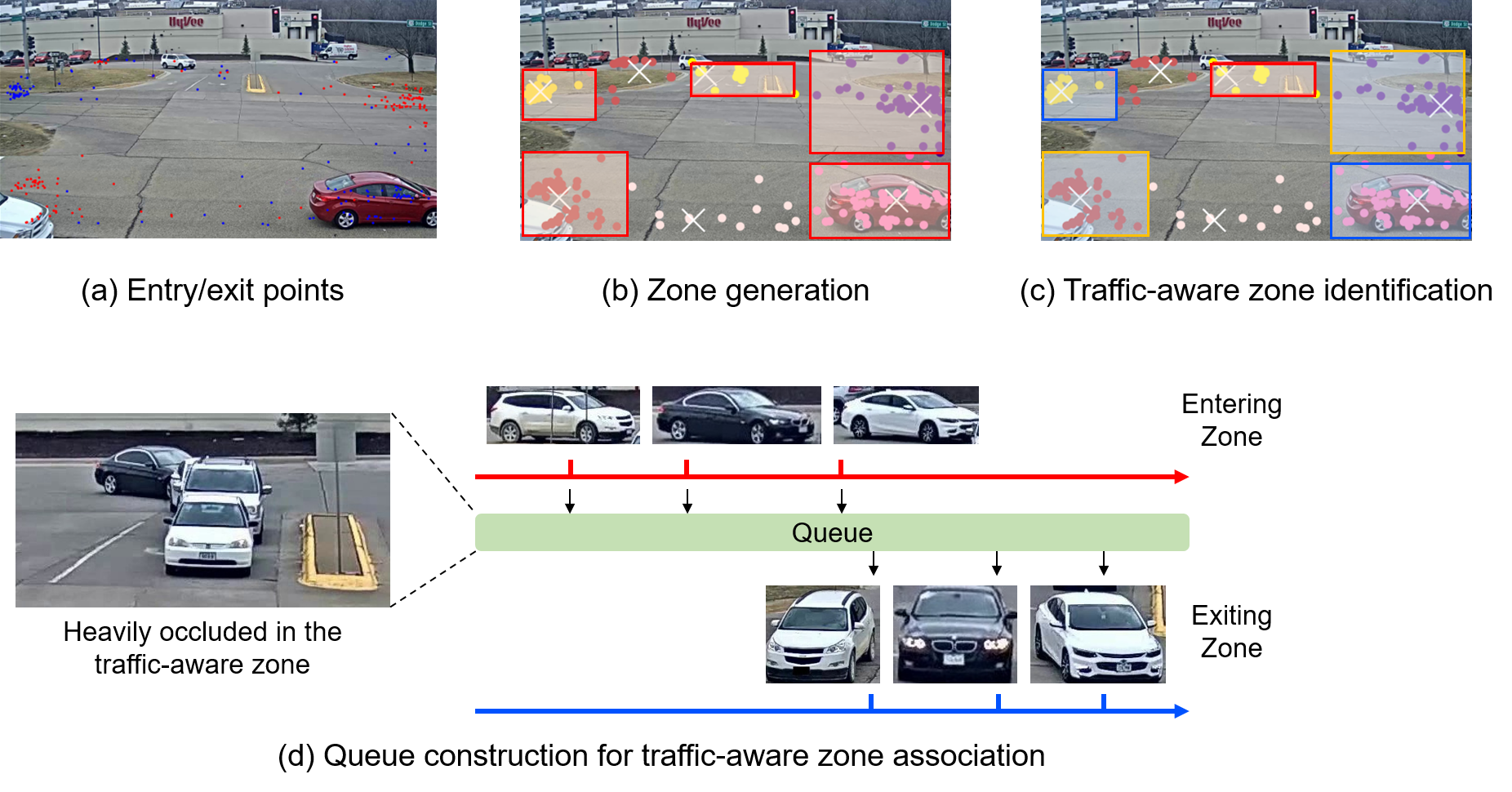}
\caption{Traffic-aware zone generation. (a) Exit/entry points of all trajectories are obtained from SCT (red: entry; blue: exit). (b) The generated zones from MeanShift. (c) The traffic-aware zones, entry zones and exit zones are identified based on traffic-aware zone density, entry zone density and exit zone density, respectively (red: traffic-aware zone; yellow: entry zone; blue: exit zone). (d) A queue for isolated trajectories is maintained to keep the ordering for following single camera ReID.}
\label{fig:traffic_aware_zone}
\end{figure}

In an MTMCT system, the first step is to perform SCT. Before we illustrate our SCT approach, we provide the definition of the SCT problem first. The input of MTMCT is $V$ videos from $V$ cameras so that we can denote the global trajectory set as $\mathcal{T} = \left\{ \Xi^{1}, \Xi^{2}, \cdots, \Xi^{V} \right\}$. Each element $\Xi^{i}$ in $\mathcal{T}$ indicates local trajectory set, which includes all trajectories in camera $i$. We then define the local trajectory set as $\Xi^{i} = \left\{ \xi^{i}_{1}, \xi^{i}_{2}, \cdots, \xi^{i}_{j} \right\}$, where $i$ is the index of camera and $j$ denotes the index of trajectory in camera $i$. Therefore, the purpose of SCT is to produce $\Xi^{i}$. 

Overall, the procedure of the proposed traffic-aware single camera tracking (TSCT) is as follows: (1) Use TNT to generate trajectories set $\Xi$ of each camera. (2) Use $\Xi$ as the input of MeanShift to generate the zones. (3) Calculate the exit density $D_x$, entry density $D_e$ and traffic-aware zone density $D_{ta}$ to classify the traffic-aware zone, exit zone and entry zone. (4) Merge the isolated trajectories in the traffic-aware zones based on appearance feature similarity and IOU for trajectory refinement. We will illustrate the intuition and the detail of each step in the following paragraphs.

The main idea of TSCT is to take advantage of traffic rules to refine the generated trajectories from SCT by merging the isolated trajectories, which are the trajectories that suddenly disappear or appear in the middle of the image. Here we use the TrackletNet Tracker (TNT) \cite{wang2018exploit} as our SCT tracker. TNT is a tracklet based graph clustering model, where each vertex in the graph is a tracklet instead of a detection. In order to establish a tracklet based graph, we need to generate tracklets as vertices. In \cite{wang2018exploit}, tracklet generation is based on detection association. First, we use metric learning to train our CNN based appearance feature extractor. Then, the tracklets are generated based on the appearance similarity and the intersection-over-union (IOU) from all detections between two consecutive frames. The tracklets are represented as the vertices of the tracklet based graph. In terms of the edges of a tracklet based graph, the edge weights are estimated by a TrackletNet, which is a Siamese neural network trained to calculate likelihood of the two tracklets being from the same object. A TrackletNet combines both temporal and spatial features as input to estimate the likelihood. After tracklet based graph construction, clustering \cite{tang2018single} is applied to merge tracklets from the same vehicle into one group.

The purpose of TSCT is to use traffic rules to improve the SCT results \cite{hsu2020traffic}. Therefore, the first step is to generate the traffic-aware zones. We notice that there are many isolated trajectories in the generated SCT results and most of them locating in the same place. It turns out that the vehicles need to wait for the stop sign or traffic light. Therefore, the window size for TNT to associate the detections results is not long enough to cover the necessary time for the traffic rules. To solve this problem, we propose an unsupervised solution to generate the traffic-aware zones, which can be exploited to reconnect these isolated trajectories to achieve trajectory refinement. In TSCT, we use MeanShift to produce these traffic-aware zones based on exit/entry point of each trajectory. Therefore, the procedure of zone generation can be divided into the following steps. First of all, we define the entry point as the first position $P_{j,f}$ and the exit point as the last position $P_{j,l}$ of the $j$th trajectory, then we apply these points as the input of MeanShift algorithm. The main idea here is to refer these points as nodes, which can be clustered to generate several clusters as zones. Since each generated cluster can be encompassed by a rectangular bounding box, which represents a zone. We can categorize these clustered zones into three types: exit zone, entry zone and traffic-aware zone. The use of exit zones and entry zones, as part of the TCLM, which will be discussed in the Section \ref{sec:clm}. On the other hand, the traffic-aware zone is used to reconnect the isolated trajectories, which are trajectories terminated or initiated in this zone. Here we use MeanShift as the clustering approach, which works by seeking the candidates for centroids within a given region. Given a candidate centroid $c_{i}$ at iteration $t$, the candidate is updated according to $c^{t+1}_{i}$= $m(c^{t}_{i})$. $N(c_{i})$ is the neighborhood of samples within a given distance around $c_{i}$ and $m$ is the mean shift vector which is calculated for each centroid that indicates towards a region of the maximum increase in the density of nodes. The following equation effectively updates a centroid to be the mean of the samples among its neighborhood:
\begin{equation}
m(c^{t}_{i})=\frac{\sum_{c_{j}\in N(c^{t}_{i})} K(c_{j}-c^{t}_{i})c_{j}}{\sum_{c_{j}\in N(c^{t}_{i})} K(c_{j}-c^{t}_{i})}, 
\end{equation}
\begin{equation}
K(c_{j}-c^{t}_{i})=exp(-\frac{||c_{j}-c^{t}_{i}||}{2\sigma^2}).
\end{equation}
Here $\sigma$ is the bandwidth of radial basis function kernel $K$, which is a parameter used to indicate the size of the region to search through. 

After MeanShift, we generate the encompassing bounding boxes for each cluster as zones in the camera and compute the entry/exit zone density to determine the type for each clustered zone. The number of nodes in each zone needs to be over a specific threshold; otherwise, the zone will be removed. Thus, the entry and exit zone densities in each zone are defined as $D_e$ and $D_x$, where 

\begin{equation}
D_e = \frac{N_{e,k}}{N_{e,k} + N_{x,k}}, \ 
D_x = \frac{N_{x,k}}{N_{e,k} + N_{x,k}}.
\end{equation}
$N_{x,k}$ and $N_{e,k}$ are the number of exit points and entry points in each zone, respectively. If the density of an entry or exit zone is higher than a threshold $\rho_e$ or $\rho_x$, this zone will be recognized as an entry or exit zone, respectively. In terms of traffic-aware zone, the traffic-aware zone density $D_{ta}$ is defined by 
\begin{equation}
D_{ta} = 1 - \frac{|N_{e,k} - N_{x,k}|}{N_{e,k} + N_{x,k}},
\end{equation}
where $D_{ta}$ needs to be above a threshold $\rho_{ta}$, then the zone will be designated as a traffic-aware zone.

\begin{equation}
Z = \left\{ \begin{array}{ll}
entry \ zone & \textrm{if $D_e > \rho_e$,}\\
exit \ zone & \textrm{if $D_x > \rho_x$,}\\
traffic-aware \ zone& \textrm{if $D_{ta} > \rho_{ta}$,}\\
don't \ care& \textrm{otherwise.}
\label{eq:zone_class}
\end{array} \right.
\end{equation}

Finally, we trained a ReID model specifically for the isolated trajectories reconnection. The TSCT follows the First-In-First-Out (FIFO) strategy to merge trajectories, i.e., the TSCT takes into account the order and temporal constraint into the tracklet grouping. An example of TSCT is illustrated in Fig.~\ref{fig:traffic_aware_zone}, where three vehicles are waiting for the trafﬁc sign so that there are three exit points in the trafﬁc-aware zone. Consequently, there are three new tracklets appearing in the trafﬁc-aware zone due to the changes of trafﬁc sign. We keep the temporal ordering of the three exit points and adopt the FIFO strategy to reconnect these three corresponding tracklets corresponding to these three exit points to the newly appeared tracklets. Speciﬁcally, we use FIFO and the order constraint (i.e., Fig.~\ref{fig:transition_time} (b)) for the single camera ReID.



\subsection{Metadata-Aided ReID}

Metadata-Aided ReID is to combine appearance embedding and metadata feature to generate the final embedding feature, then we can use the final embedding feature for ICT. In this section, we will illustrate the training procedures of our appearance ReID model and metadata feature.

After TNT tracking, we obtain the tracking results for each camera, which are the input data of ReID. In other words, we can take advantage of a sequence of images, instead of a single image, since video-based ReID can achieve better performance than image-based ReID. For frame-level feature extraction, we adopt the ResNet-50 \cite{he2016deep} network pre-trained on ImageNet as our feature extractor, and the appearance feature of an object is obtained from the 2048-dim fully-connected layer. After frame-level feature extraction, temporal information can be further taken advantage of to establish a more discriminative feature. To this end, we use temporal attention (TA) mechanism \cite{gao2018revisiting} to perform a weighted average the frame-level feature and create a clip-level feature. The main idea is that some frames of the object might be highly occluded by other objects, then we want to lower the weight of these frames. We set a specific number of frames to do the TA to generate the clip-level feature. Finally, we add another average pooling layer for these clip-level feature $f_c$ to generate the final tracklet-level feature. 

There are two convolutional networks, which are spatial convolutional network and temporal convolutional network, used in the TA model. Assume the clip size is $c$, the spatial convolutional network is a 2D convolution operation to produce $c \times 256$-dim feature vectors, then we apply 1D temporal convolutional network with kernel size 3 to generate an attention vector for weighting the frame-level feature so that the clip-level feature $f_c$ can be created.

For the network training, we adopt the metric learning strategy by using the batch sample (BS) \cite{kumar2019vehicle} in the triplet generation. In terms of the loss function, there are two types of loss to be jointly optimized, which are BS triplet loss and cross-entropy loss. Thus, the final loss function of our network is a combined loss, described as follows:

\begin{equation}
\mathcal{L}_{total} = \lambda_1 \mathcal{L}_{BStri} + \lambda_2 \mathcal{L}_{Xent}.
\end{equation}
First, the triplet loss is to minimize the feature distance of the same identity and maximize the feature distance of different identity pairs \cite{hermans2017defense}. In this paper, we adopt BS triplet loss to calculate triplet loss in a minibatch $B$, which can be defined as,

\begin{equation}
\mathcal{L}_{BStri} (\theta; \xi) = \sum_{b} \sum_{a \in B} l_{triplet}(a),
\end{equation}
where
\begin{equation}
l_{triplet}(a) = \left[ m + \sum_{p \in P(a)} w_p D_{ap} - \sum_{n \in N(a)} w_n D_{an}\right] _+.
\end{equation}
$m$ denotes the margin, $D_{ap}$ and $D_{an}$ represent the distances between the anchor sample $a$ to the positive instance and negative instance, respectively. $w_{p}$ and $w_{n}$ are the weights of positive and negative instances, which are defined as follows,

\begin{equation}
\begin{aligned}
w_p = P(w_p == \text{multinomial}_{x \in P(a)} \{D_{ax} \}),\\
w_n = P(w_n == \text{multinomial}_{x \in N(a)} \{D_{ax} \}),
\end{aligned}
\end{equation}
where $P(a)$ and $N(a)$ are positive and negative instances, respectively.

The cross-entropy (Xent) loss [24] in the training is defined as follows,
\begin{equation}
\mathcal{L}_{Xent} = - \frac{1}{N} \sum_{j=1}^{N} y_{j}\log(\hat{y_{j}}),
\end{equation}
where $\hat{y_{j}}$ is the estimated probability of the probe object that belongs to object $j$, and $y_{j}$ denotes the ground truth vector while $N$ denotes the number of identities in training data.

\begin{figure}[t]
\begin{center}
\includegraphics[width=\linewidth]{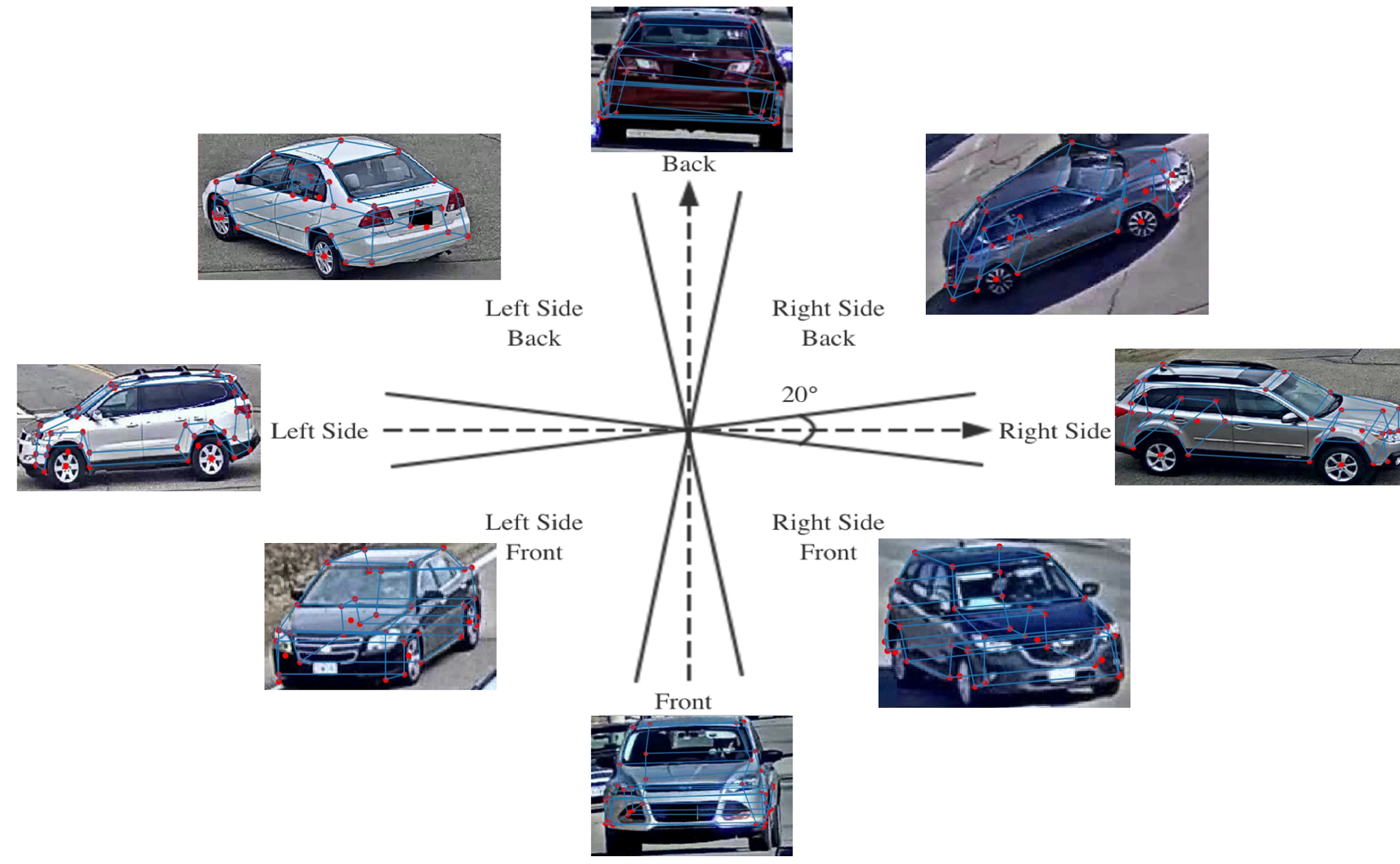}
\end{center}
 \caption{Orientation view perspectives for data augmentation.}
\label{fig:ori}
\end{figure}

\begin{figure}[t]
\begin{center}
\includegraphics[width=0.8\linewidth]{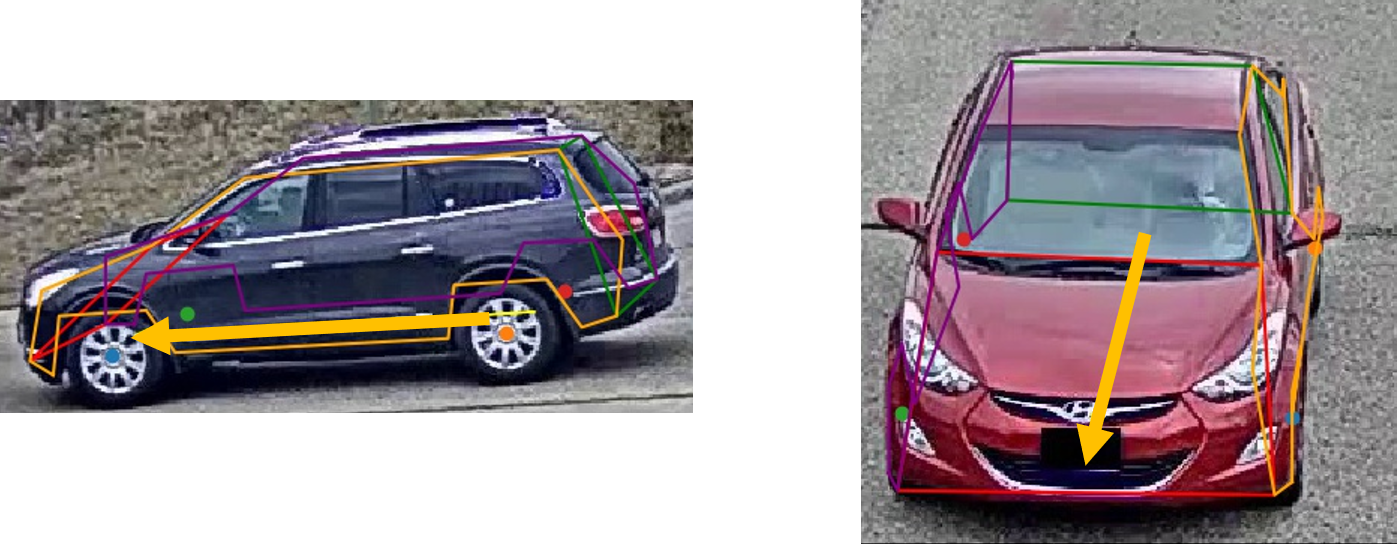}
\end{center}
 \caption{Examples of vehicle keypoints detection and visibility estimation. The red, green, orange and purple outlines represent the front, back, left side and right side of vehicles, respectively. Notice that the yellow arrow represents the driving direction $\vec{r}$, and the data augmentation is based on the driving direction angle of $\vec{r}$ to sample different car orientations for training.}
\label{fig:kp}
\end{figure}

\textbf{Metadata classification.} In the proposed Metadata-Aided ReID, vehicle type, brand and color are considered as vehicle metadata attributes. We adopt a 29-layer Light CNN framework \cite{wu2018light}, including small kernel sizes of convolution layers, network-in-network layers and residual blocks. This Light CNN architecture is used to reduce the parameter space to improve the performance of metadata classification. The max-feature-map (MFM) operation is an extension of maxout activation, which combines two feature maps and outputs element-wise maximum value. Finally, we use the output layer of this Light CNN framework as the metadata feature for vehicle ReID. Assume the metadata feature of trajectory $j$ can be represented as $\xi_j=\left\{a_{1}, \cdots, a_{n} \right\}$, and $a_{i}$ denotes the output of the Light CNN for the $i$th frame of $\xi_j$ (i.e., the probability distribution of metadata classification). Thus, the final metadata feature ${\mathcal{M}(\xi_j)}$ of a trajectory $\xi_j$ is defined as follows: 
\begin{equation}
\begin{aligned}
{\mathcal{M}(\xi_j)={\frac {1}{n}}\sum _{i=1}^{n}a_{i}}.
\end{aligned}
\end{equation}

In this work, we utilize data augmentation to achieve better results of metadata classification. First of all, we use the car pose algorithm to produce the 36 keypoints for each vehicle \cite{ansari2018earth}. Based on the 36 vehicle keypoints, we can mark the four wheels as $P_{front,left}$, $P_{front,right}$, $P_{back,left}$ and $P_{back,right}$. Then, the direction of a vehicle can be represented by a vector that is pointing from the center of the back axle to the center of the front axle,

\begin{equation}
\vec{r}=((P_{back,left}+P_{back,right})/2,(P_{front,left}+P_{front,right})/2).
\label{eq:angle}
\end{equation}
Therefore, we can define the direction of a vehicle through the angle of $\vec{r}$ with respect to the horizontal right-sided view perspective, i.e., $0^{\circ}$. Fig.~\ref{fig:ori} shows that the 2D space is split into 8 regions which are calculated by 
\begin{equation}
\mathcal{R} = \left\{ \begin{array}{ll}
\ [\theta - \phi , \theta + \phi)& \textrm{if $\theta \in \{0^{\circ},90^{\circ},180^{\circ},270^{\circ}\}$}\\
\ [\theta , \theta + \omega) & \textrm{if $\theta \in \{10^{\circ},100^{\circ},190^{\circ},280^{\circ}\}$}.\\
\end{array} \right.
\label{eq:regions}
\end{equation}
Here the $\phi$ and $\omega$ are set as $10^{\circ}$ and $70^{\circ}$, respectively.
Fig.~\ref{fig:kp} shows an example of using vehicle keypoints to estimate the vehicle directions and visibility. In the data augmentation, we cluster all the images by these 8 different vehicle directions and enlarge the input size to be 512 $\times$ 512, the aspect ratio is retained by zero paddings. We believe the vehicle directions including the semantic meaning based on the visible surfaces can be exploited to diversify the variation of training data. We also slightly change the architecture of\cite{wu2018light} by adding one network-in-network layer and extend the dimensions of the fully-connected layer from 256 to 2048. The classes of metadata and implementation details are illustrated in Section \ref{sec:results}. 

\subsection{Trajectory-based Camera Link Model}\label{sec:clm}

\begin{figure}[t]
\centering
\includegraphics[width=0.88\linewidth]{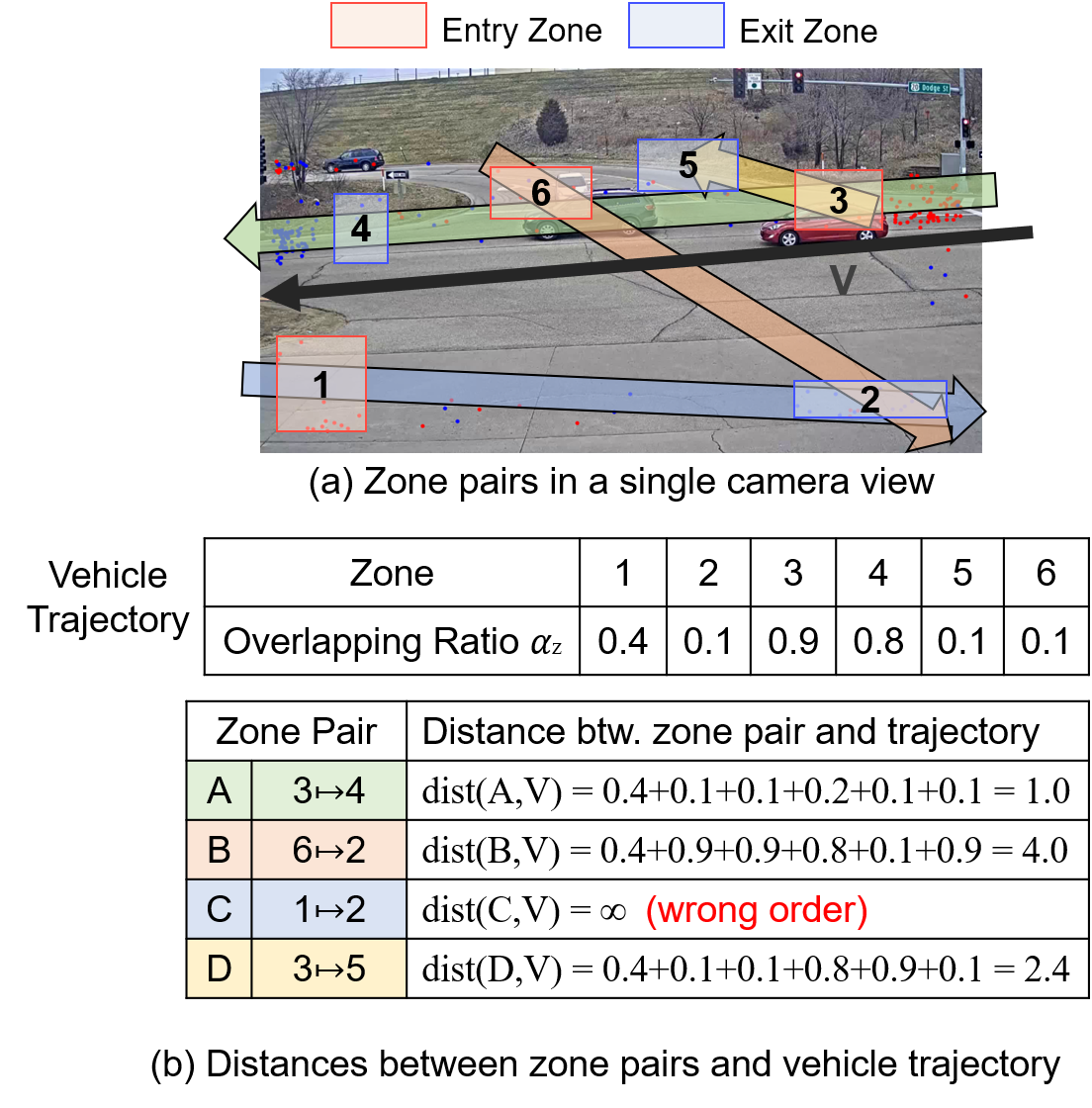}
\caption{An example of distance calculation between a vehicle trajectory (black) with respect to each entry-exit zone pair. Assume that there are four zone pairs (A, B, C, D) which are two entry and two exit zones in a single camera. If there is a vehicle passing through the four zones with the overlapping area ratio $\alpha_z$ in the upper table in (b), the distances can be estimated by Eq.~(\ref{eq:dist_vt_pairs}). Then, this vehicle trajectory can be classified into the zone pair with the smallest distance. In this case, it is zone pair A.}
\label{fig:zone_pairs}
\end{figure}

\begin{figure}
\centering
\includegraphics[width=0.8\linewidth]{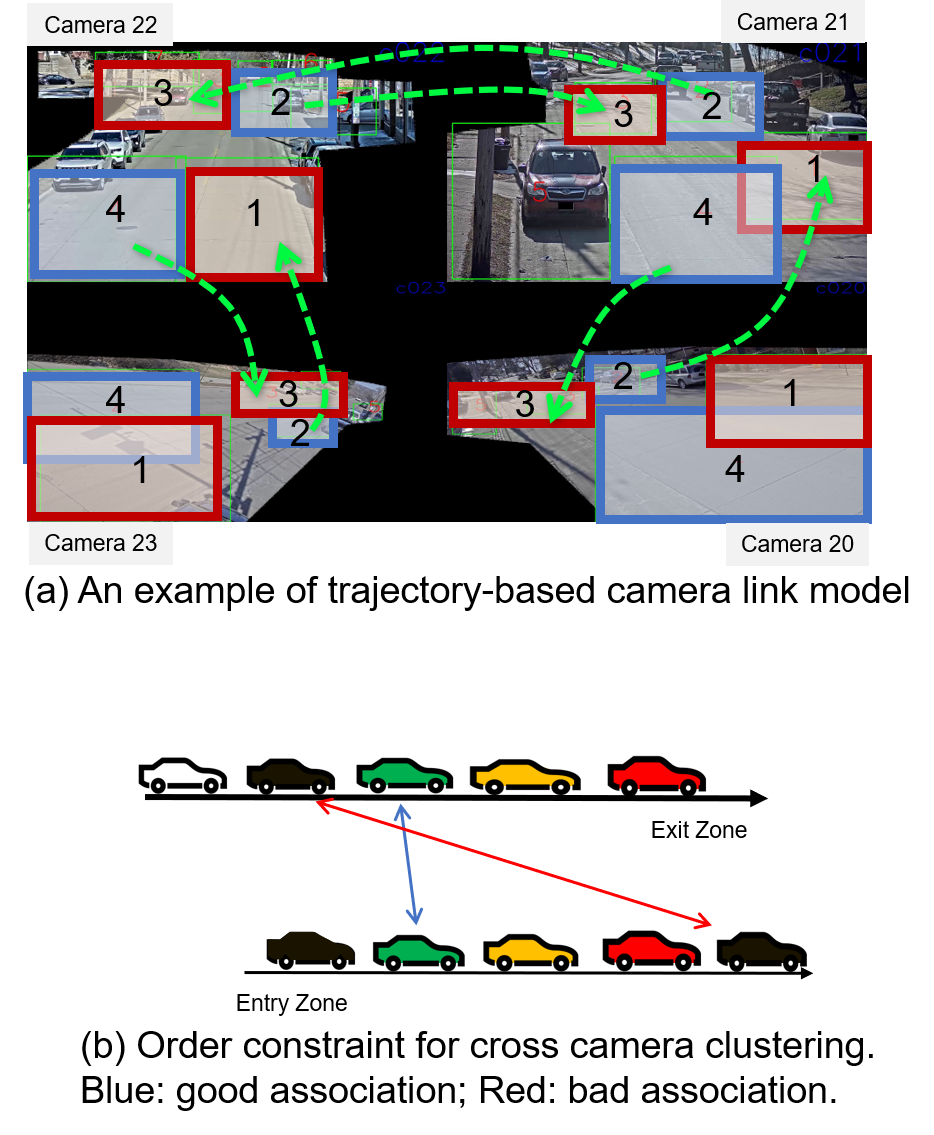}
\caption{Illustration for (a) trajectory-based camera link model, and (b) the order constraint. There are four cameras with overlapping FoVs: camera 20, camera 21, camera 22 and camera 23. The exit/entry zones are denoted as blue/red bounding boxes, respectively. In (a), the exit zone in Camera 21 (blue bounding box) and the entry zone in Camera 22 (red bounding box) indicate the camera connectivity. In this scenario, the vehicles exiting from the camera 21 will appear in the Camera 22 immediately, similarly for camera 22 to camera 23 and camera 20 to camera 21. Take the connectivity of camera 21 and camera 22 as an example, there is a transition camera link ${\L}$ including $C^{21}$ and $C^{22}$, where $C^{21}=\{P_1^{21},P_2^{21}\}$ and $C^{22}=\{P_1^{22},P_2^{22}\}$ have two zone pairs, respectively. Therefore, $z_{s}=2 \in P_1^{21} $ and $z_{d}=3 \in P_2^{22}$ are one transition zone pair while $P_1^{21}=\{1,2\}$ and $P_2^{22}=\{3,4\}$. Another transition zone pair are $z_{s}=2 \in P_1^{22} $ and $z_{d}=3 \in P_2^{21}$ while $P_1^{22}=\{1,2\}$ and $P_2^{21}=\{3,4\}$.}
\label{fig:transition_time}
\end{figure}

Camera link model (CLM) is to exploit the camera topological configuration to generate the temporal constraint so as to improve the performance of vehicle ReID. The definition of the CLM is to connect adjacent cameras that are directly connected to each other. If no existing cameras need to be passed between the two cameras, the two cameras form a camera link. 
The link between two directly-connected cameras actually only connects one zone each in these two cameras. There may be multiple entry/exit zones within a camera's view. 
Due to road structures and traffic rules, the motion of the vehicles follows certain driving patterns. Thus, we group the SCT results into limited numbers of trajectories, then exploit the limited numbers of trajectories to establish trajectory-based camera link model (TCLM). It means that the CLM can be more specific into trajectory level to generate a more accurate transition time distribution of each camera link.

In order to generate TCLM, we need to distinguish trajectories that are determined by driving patterns. We define the zones of each camera so that the zone pair can be used to uniquely describe a trajectory (Fig.~\ref{fig:zone_pairs}). The zones can be the intersection areas, the turning points of a road or the enter/exit areas of the camera’s field-of-view. On the other hand, these zone pairs also need to be labeled to complete TCLM. In our work, we generate all possible zones and use the training data of MTMCT to automatically generate the camera links instead of human labeling. In most of the cases, the straight and the right-turn trajectories can be described using different zone pairs which are gone through by the passing vehicles. Sometimes the trajectory may not go through the corresponding prespecified zone pair perfectly due to the viewing angle of the camera. In this case, measuring the distance between a tracked vehicle and a zone pair is necessary. The distance can be calculated as,
\begin{equation}
\text{dist}(P, V) = \sum_{z \in P \cup V} |\mathds{1}(z \in P) - \alpha_z|,
\label{eq:dist_vt_pairs}
\end{equation}
where $P$ denotes the zone pair and $V$ is the actual zones gone through by the tracked vehicle. $\alpha_z$ represents the overlapping ratio of the vehicle to zone $z$ (i.e., the overlapping area divided by the vehicle bounding box area). Furthermore, the order of the zones in the zone pair and the tracked vehicle are also considered. Once the order in the tracked vehicle conflicts with the zone pair, the distance between a tracked vehicle and a trajectory is set to infinity. Finally, the closest trajectory is assigned by comparing the tracked vehicle with all the possible zone pairs within the camera.

The last step of TCLM generation is to estimate the transition time of camera links. After getting the trajectories and camera links, the transition of each camera link can be defined as ${\L} = (C^{s}, C^{d})$, where $C^{s} = \{ P^{s}_{i}\}_{i=1}^m$ is the zone pair set in the source camera and $C^{d} = \{P^{d}_{j}\}_{j=1}^n$ is the zone pair set in the destination camera. Each camera link can have more than one transition due to the bidirectional traffic. An example is shown in Fig.~\ref{fig:transition_time} to illustrate the concept of TCLM. We first define the transition zone pair $z_{s}$ and $z_{d}$, such that $z_{s} \in P^s_{i}$ ($\forall P^s_{i} \in C^s$) and $z_{d} \in P^d_{j}$ ($\forall P^d_{j} \in C^d$). Then, the transition can be imposed with the temporal constraint for both $C^{s}$ and $C^{d}$. Given a camera link of a vehicle trajectory from $P^s$ to $P^d$ ( i.e., from the source camera to the destination camera), the transition time is defined as 
\begin{equation}
\Delta t = t^s - t^d,
\end{equation}
where $t^s$ and $t^d$ represent the time the tracks passing $z_s$ and $z_d$ respectively. Then, we can obtain a time window $(\Delta t_{\text{min}}, \Delta t_{\text{max}})$ for each camera link ${\L}$, so that only the tracked vehicle pairs with transition time within the time window are considered as valid. Thus, the search space of the ReID can be reduced by an appropriate time window.




\subsection{Global Trajectory Generation}

\begin{figure}[t]
\begin{center}
 \includegraphics[width=1.0\linewidth]{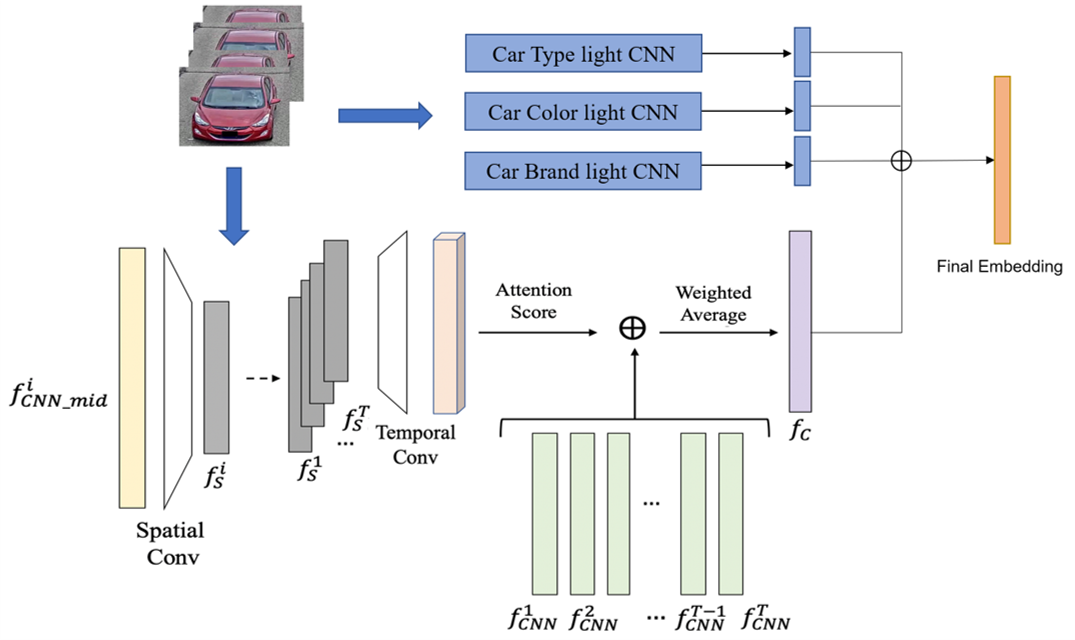}
\end{center}
 \caption{Illustration for the final embedding features. The appearance feature is generated by a temporal attention architecture based on ResNet-50 as the backbone. Three metadata features are generated by three separately trained Light CNN networks. The final representation for global trajectory generation is formed by concatenating all these four features.}
\label{fig:global-trajectory-generation}
\end{figure}

\begin{figure}[t]
\centering
\includegraphics[width=0.85\linewidth]{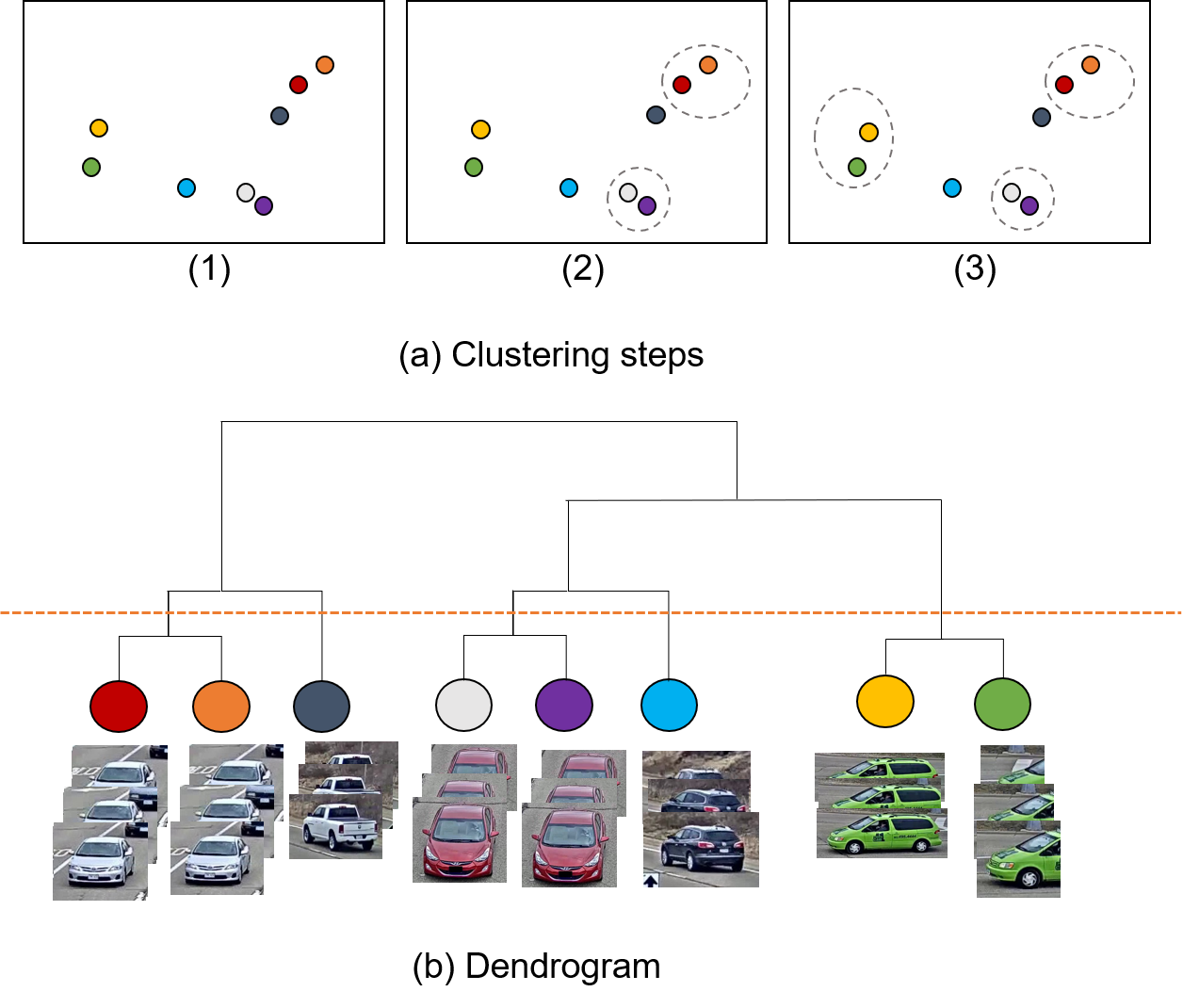}
\caption{The procedure of hierarchical clustering: (a) the clustering steps, (b) the dendrogram of clusters with different threshold levels.}
\label{fig:hc}
\end{figure}

\begin{algorithm}[t]
\SetAlgoLined
\SetKwInOut{Input}{Input}
\SetKwInOut{Output}{Output}
\SetKw{KwFrom}{from}
\SetKw{KwAnd}{and}
\Input{Trajectories set $\mathcal{T}$ from all cameras.}
\Output{Global ID for all trajectories within all cameras.}

 Initialize distance matrix $\mathbf{M}$ between each two trajectories $\mathbf{M} \gets \infty$\;
 \For{trajectory $\xi_i$ \KwFrom $\mathcal{T}$}{
\For{trajectory $\xi_j$ \KwFrom $\mathcal{T}$}{
 Calculate $\mathbf{M}_{i,j}$ by Eq.~(\ref{eq:dist}).
}
 }
 Flatten and sort the upper triangular part of $\mathbf{M}$ in ascending order: $F \gets sort([\mathbf{M}_{i,j}]_{i<j})$\;
 $iter \gets 0$, $\delta \gets$ distance threshold\;
 \While{$iter <$ \# of iterations}{
\For{$m_{i,j}$ \KwFrom $F$}{
\eIf{$m_{i,j} < \delta$ \KwAnd valid order constraint for $\xi_i$, $\xi_j$}{
 Assign the same global ID to $\xi_i$ and $\xi_j$\;
 }{
 $m_{i,j} \gets \infty$\;
 }
}
$iter ++$\;
 }
 \caption{Hierarchical Clustering}
\label{code:whole_hierarchical}
\end{algorithm}

MTMCT can be referred as a correlation clustering problem to generate the global trajectory, which can be defined as the Binary Integer Program (BIP).
Here, we describe the BIP as follows. Assume that there is a weighted graph $\mathcal{G} = (\mathcal{V},\mathcal{E},\mathcal{W})$, where $\mathcal{V}$ represents the single camera trajectory node set, the weight $\mathcal{W}$ of the edge $\mathcal{E}$ represents the corresponding correlation between the nodes. 
\begin{equation}
\begin{aligned}
X^* &= \arg \max_{\{x_{i,j}\}} \sum_{(i,j) \in \mathcal{E}} w_{i,j} x_{i,j}, \\
\text{s.t.,} \ & x_{i,j} + x_{j,k} \leq 1 + x_{i,k}, \ \ \forall i,j,k \in \mathcal{V}.
\end{aligned}
\label{eq:bip}
\end{equation}
The set $X$ represents the set of all possible combinations of assignments to the binary variables $x_{i,j}$. Here $w_{i,j}$ means the weight between two trajectories $i, j$, which can be obtained by calculating the distance of the two trajectories $i, j$. By rewarding edges that connect the same vehicle's multi-camera trajectories and penalizing edges that link to the different vehicles, we can maximize $X^*$. For example, the $x_{i,j}$ of two multi-camera trajectories $i, j$ should be assigned $1$ if trajectories $i, j$ indicate the same vehicle identity. Meanwhile, the constraints in Eq.~(\ref{eq:bip}) are used to enforce the transitivity in the solution. 

Then, we can use the final embedding feature $\mathbf{f}(\xi_i)=\mathcal{A}(\xi_i)\oplus\mathcal{M}(\xi_i)$, TCLM and hierarchical clustering algorithm to produce the global IDs for MTMCT. The illustration of generating the final embedding is shown in Fig.~\ref{fig:global-trajectory-generation}, and the procedure of hierarchical clustering is in Fig.~\ref{fig:hc} and Alg. \ref{code:whole_hierarchical}.
$\mathcal{A}(\xi_i)$ and $\mathcal{M}(\xi_i)$ indicate the ReID feature and metadata feature, respectively. $\oplus$ denotes the concatenation operator. We can thus define a distance matrix 
\begin{equation}
\begin{aligned}
\textbf{M}=\left[
\begin{array}{ccc}
 \textbf{M}_{11} & \cdots & \textbf{M}_{1N} \\
 \vdots & \ddots & \vdots \\
 \textbf{M}_{N1} & \cdots & \textbf{M}_{NN}
\end{array}
\right],
\end{aligned}
\end{equation}
where 
\begin{equation}
\textbf{M}_{i,j} = \left\{ \begin{array}{ll}
\text{dist}(\mathbf{f}(\xi_i), \mathbf{f}(\xi_j)) & \textrm{if valid camera link constraint,}\\
0 & \textrm{otherwise,}
\label{eq:dist}
\end{array} \right.
\end{equation}
which represents all the distance between any two trajectories from two different cameras. Here,
\begin{equation}
\begin{aligned}
\text{dist}(\mathbf{f}(\xi_i),\mathbf{f}(\xi_j))&=\text{dist}(\mathcal{A}(\xi_i)\oplus\mathcal{M}(\xi_i),\mathcal{A}(\xi_j)\oplus\mathcal{M}(\xi_j))\\
 &= \left\|\mathcal{A}(\xi_i)\oplus\mathcal{M}(\xi_i)-\mathcal{A}(\xi_j)\oplus\mathcal{M}(\xi_j)\right\|_2. 
\end{aligned}
\end{equation}
Moreover, the time window $(\Delta t_{\text{min}}, \Delta t_{\text{max}})$ of each camera link is generated as the CLM constraint. Since the search space of ReID is reduced by the CLM constraint, the Rank-1 accuracy can be improved. By using hierarchical clustering, we greedily select the smallest pair-wise distance to merge the tracked vehicles cross cameras. 

Furthermore, the orders between different tracked vehicles can be used as an additional constraint to further reduce the search space of the ReID. Due to the traffic rule or the road condition, the orders of vehicles should be almost the same. Take Fig.~\ref{fig:transition_time}(b) as an example, we will remove the pairs which conflict with previously matched pairs. The process will be repeated until there is no valid transition pair or the minimum distance is larger than a threshold. Given two vehicle trajectories $\xi_{src_1}$, $\xi_{src_2}$ in source camera and two vehicle trajectories $\xi_{dst_1}$, $\xi_{dst_2}$ in destination camera, we define
\begin{equation}
\begin{aligned}
\text{sign}(\xi_{src_1} - \xi_{src_2}) = \text{sign}(\xi_{dst_1} - \xi_{dst_2}),
\end{aligned}
\end{equation}
i.e., the orders of tracked vehicles in source and destination camera should remain the same. In this case, the search space can be further reduced. 


\section{Experiments}\label{sec:results}

\begin{figure}[t]
\centering
\includegraphics[width=\linewidth]{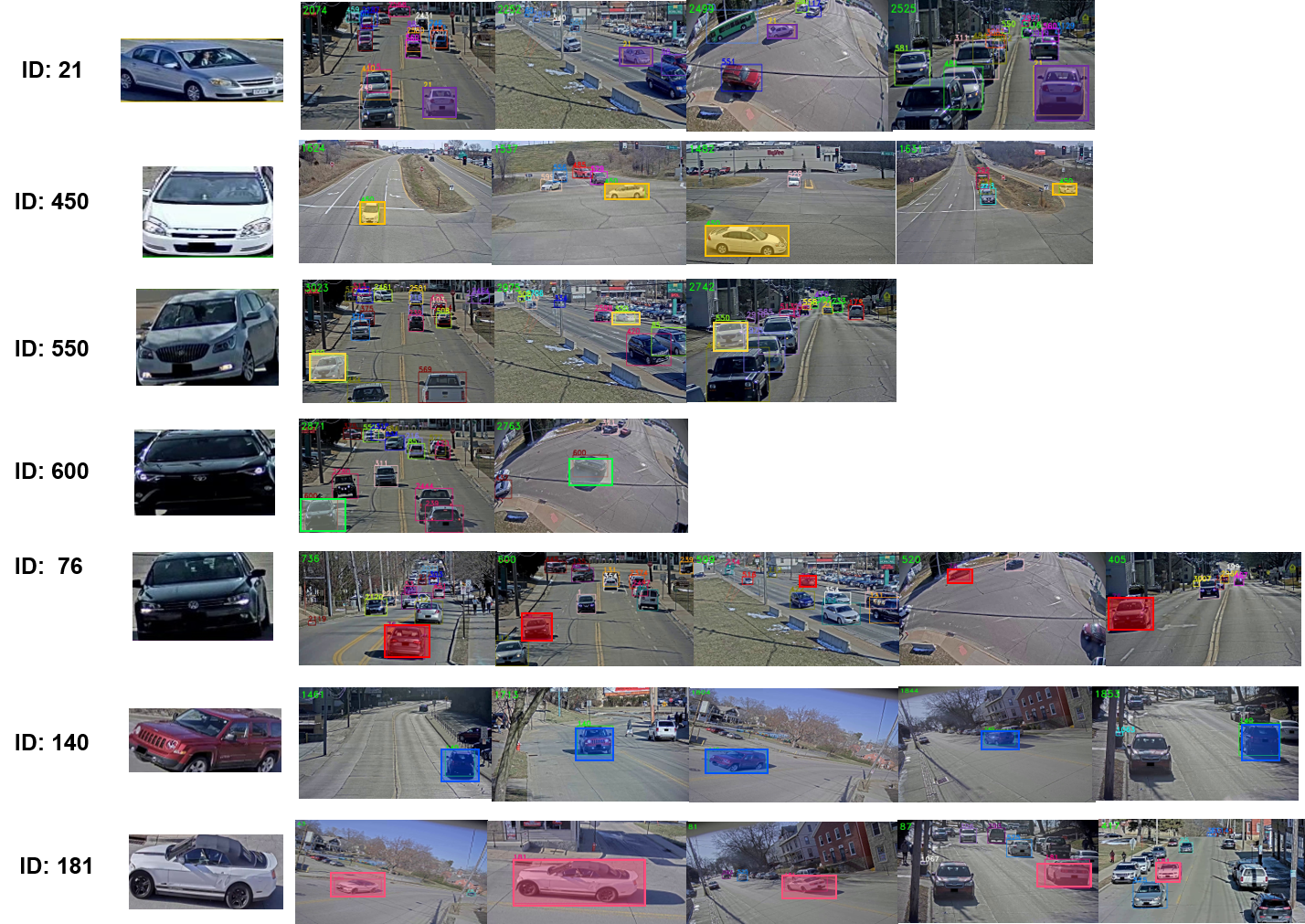}
\caption{Qualitative results of the proposed method on CityFlow dataset.}
\label{fig:qual_results}
\end{figure}


\subsection{CityFlow MTMCT Dataset and Implementation Details}
In this section, we evaluate the proposed method using a benchmark city-scale MTMCT dataset CityFlow \cite{tang2019cityflow} and compare it with state-of-the-art MTMCT methods. To the best of our knowledge, CifyFlow is the only one city-scale multiple camera vehicle tracking dataset, where the TCLM is established by the training data, and the cameras in the testing data are overlapped with the training data. There are 3.25 hours of videos in CityFlow, which are collected from 40 cameras across 10 intersections spanning over 2.5 miles of distance coverage in a mid-sized U.S. city. There are many different types of road scenarios in CityFlow (e.g., intersections, stretches of roadways, and highways). Moreover, CityFlow contains 229,680 bounding boxes for 666 vehicles, whose license plates are all obscured due to privacy concerns.

Our experimental environment is set as follows: the SCT is based on Python: 3.5.4 and Tensorflow: 1.4.0; the ReID model is realized using the PyTorch 0.3.1, Torchvision 0.2.0 and Python 2.7. One NVIDIA Quadro GV100 32GB GPU is used for our experiments. As for the data used for training, the details are listed as follows. (1) For SCT, we train the TNT using the training data of AI City Challenge 2018 dataset. (2) For ReID, the training data consists of two parts, i.e., the cropped vehicles from the training data of CityFlow and the training data of CityFlow-ReID dataset. (3) In terms of metadata classifier, we train on the training data of CityFlow-ReID dataset. (4) The TCLM is trained on the training data of CityFlow dataset.

Some implementation details of the proposed methods, i.e., SCT and MA-ReID, are detailed here. First of all, as the baseline for TSCT we use the TNT, which is pretrained on the AI City Challenge 2018 dataset [47] that contains over 3.3K vehicles. The dimensionality of the appearance feature of TNT is 512, the time window of tracklet generation is set as 64 frames, and the batch size for training TNT is 32. We select Adam as the optimizer for TNT and the learning rate is from $10^{-3}$ to $10^{-5}$ with the weight decay 10 times in every 2000 steps. 

Next, the parameter of MeanShift is the bandwidth of the radial basis function kernel, which is set as 250 for generating the appropriate area size of zones. According to our experiments, the area of the traffic-aware zone is supposed to be at least 150\% of the average area of the vehicle bounding box. Moreover, the traffic-aware density $\rho_{ta}$ of each zone is set as 0.8 to ensure this zone is a traffic-aware zone, otherwise the traffic-aware zone is not reliable and we rather do nothing. Furthermore, $\rho_e$ and $\rho_x$ are also set as 0.8 in our experiments. After generating the zones, we need to calculate the overlapping area of the bounding boxes of the vehicles and the traffic-aware zone to locate the vehicles passing through the traffic-aware zone. 
Then, the bounding box of the last position of a tracklet will be inserted into a queue. Once there are some new trajectories suddenly appear in the traffic-aware zone, we calculate the IOU of the bounding box of the first position and the first bounding box popping from the queue. Due to the traffic jam or the queuing of vehicles, the IOU can sometimes be very small. We thus set the IOU threshold very low so as to merge these isolated trajectories. The appearance feature is not reliable in the traffic-aware zone due to serious occlusions, therefore the First-In-First-Out (FIFO) and IOU constraints are more reliable than appearance similarity. Thus, the appearance similarity threshold between isolated trajectories is set as 0.4 for isolated trajectories reconnection. This value is not sensitive because FIFO strategy and IOU constraint have imposed high restriction for isolated trajectory reconnection. The SCT results are similar from similarity score 0.1 to 0.4, and there is no isolated trajectory pair matching for similarity score over 0.5.

In terms of MA-ReID, there are two architectures used for appearance feature extractor and metadata classifier, respectively. In appearance feature extraction, the temporal attention is applied for training the ReID model, and the trained $4$-dim attention vector is used to generate the clip-level embedding based on average pooling to obtain the trajectory-level embedding. We train the ReID model for 800 epochs with batch size 32, the learning rate and weight decay are $3 \times 10^{-4}$ and $5 \times 10^{-4}$, respectively. We adopt the ResNet-50 as the backbone network for both SCT ReID and ICT ReID, and the loss function is a combination of batch-sample (BS) loss and cross-entropy (Xent) loss. In our experiments, we resize the input image size as $224 \times 224$ and ResNet-50 is pre-trained on ImageNet. 

Furthermore, the metadata classifier is trained on the training data of CityFlow ReID dataset, where 3204 low-resolution images are excluded, then the rest of the images are randomly split to Train/Validation/Test with ratio 0.7/0.1/0.2. In our experiments, we label the metadata of vehicle in the following categories: 1) Type: sedan, suv, minivan, pickup truck, hatchback and truck; 2) Brand: Dodge, Ford, Chevrolet, GMC, Honda, Chrysler, Jeep, Hyundai, Subaru, Toyota, Buick, KIA, Nissan, Volkswagen, Oldsmobile, BMW, Cadillac, Volvo, Pontiac, Mercury, Lexus, Saturn, Benz, Mazda, Scion, Mini, Lincoln, Audi, Mitsubishi and others; 3) Color: black, white, red, grey, silver, gold, blue, green and yellow. Table~\ref{tab:res0} shows that our classification accuracy in all the three metadata are over 88\%. Therefore, the MTMCT can benefit from the accurate metadata information.

\begin{table}[t]
\centering
\caption{Metadata classification results on CityFlow.}
\begin{tabular}{l|c}
\hline
Metadata& Accuracy \\
\hline
Type& 0.9483\\
Brand& 0.8801\\
Color& 0.9400\\
\hline
\end{tabular}
\label{tab:res0}
\end{table}

In MTMCT, the IDF1 score \cite{ristani2016performance} is wildly used as evaluation metrics. Thus, IDF1 is adopted to rank the performance for each participant in the CityFlow dataset, which means to calculate the ratio of correctly identified vehicles over the average number of ground-truth and predicted vehicles. The definition of IDF1 is shown as follows:
\begin{equation}
\begin{aligned}
IDP &= \frac{IDTP}{IDTP+IDFP}, \\
IDR &= \frac{IDTP}{IDTP+IDFN}, \\
IDF1 &= \frac{2IDTP}{2IDTP+IDFP+IDFN},
\end{aligned}
\end{equation}
where IDFN, IDTN and IDTP are defined as follows
\begin{equation}
\begin{aligned}
IDFN &= \sum_{\tau} \sum_{t \in T_\tau} m(\tau, \gamma_{m}(\tau), t, \Delta), \\
IDFP &= \sum_{\gamma} \sum_{t \in T_\gamma} m(\tau_{m}(\gamma), \gamma, t, \Delta), \\
IDTP &= \sum_{\tau} \text{len}(\tau)- IDFN = \sum_{\gamma} \text{len}(\gamma) - IDFP, 
\end{aligned}
\end{equation}
where the ground truth trajectory is denoted as $\tau$, $\gamma_m (\tau)$ represents the best matches of the computed trajectory for $\tau$; $\gamma$ is the computed trajectory; $\tau_m (\gamma)$ denotes the best match of ground truth trajectory for $\gamma$; $t$ indicates the frame index; $\Delta$ is the IOU threshold that determines if computed bounding box matches the ground truth bounding box (here we set $\Delta=0.5$); $m(\cdot)$ represents a mismatch function which is set as $1$ if there is a mismatch at $t$; otherwise, $m(\cdot)$ is set as $0$.

\begin{table}[t]
\centering
\caption{MTMCT results comparison on CityFlow dataset.}
\vspace{0.5em}
\begin{tabular}{l c}
\hline
Methods& IDF1 \\
\hline
MOANA+BA \cite{tang2019cityflow}& 0.3950\\
DeepSORT+BS \cite{tang2019cityflow}& 0.4140\\
TC+BA \cite{tang2019cityflow}& 0.4630\\
ZeroOne \cite{tan2019multi} & 0.5987 \\
DeepCC \cite{ristani2018features} & 0.5660\\
LAAM \cite{hou2019locality1} & 0.6300\\
ANU \cite{hou2019locality}& 0.6519 \\
TrafficBrain \cite{he2019multi} & 0.6653 \\
DDashcam \cite{li2019spatio} & 0.6865 \\
UWIPL \cite{hsu2019multi} & 0.7059 \\
\hline
Ours(TSCT+TA)& 0.7493 \\
\textbf{Ours(TSCT+TA+META)}& \textbf{0.7677} \\
\hline
\end{tabular}
\label{tab:res1}
\end{table}

\begin{table}[h]
\small
\caption{The MTMCT performance for different dimensions of the proposed method on CityFlow dataset.} 
\begin{center}
\vspace{0.3em}
\resizebox{70mm}{7mm}{
\begin{tabular}{c|c c c}
\hline 
Appearance Dimension & IDF1 & IDP & IDR\\
\hline 
512 & 0.6872& 0.6641 & 0.7120 \\
1024 & 0.7241& 0.6928 & \textbf{0.7525} \\
2048 &\textbf{0.7677}& \textbf{0.8276} & 0.7159 \\
\hline 
\end{tabular}
}
\end{center}
\label{tab:res5}
\end{table}

\begin{table}[h]
\small
\caption{The FLOPs of the proposed method.}
\begin{center}
\vspace{0.3em}
\resizebox{40mm}{6mm}{
\begin{tabular}{c|c}
\hline 
Total params & 49,199,169\\
 \hline
Total memory & 109.68MB \\
\hline 
Total FLOPs & 4.14G FLOPs \\
\hline 
\end{tabular}
}
\end{center}
\label{tab:res4}
\end{table}

\subsection{System Performance}
According to \cite{hsu2019multi,li2019spatio,he2019multi}, the spatio-temporal information can improve the performance of MTMCT. In this work, the TCLM is utilized to take advantage of spatio-temporal information to achieve the best performance. On the other hand, some systems merely use the data association graph for MTMCT. Table~\ref{tab:res1} shows that our performance can achieve the state-of-the-art on the CityFlow. Meanwhile, we compare our method with locality aware appearance metric (LAAM) \cite{hou2019locality1}, which is the state-of-the-art approach on the DukeMTMC dataset. Although DukeMTMC dataset is no longer available for MTMCT, we still can compare the performance of the proposed method with the state-of-the-art approach from DukeMTMC dataset. LAAM improves DeepCC \cite{ristani2018features} by using training the model on intra-camera and inter-camera metrics. The temporal windows of LAAM for SCT and ICT of are 500 frames and 2400 frames; they also use ResNet-50 pre-trained on ImageNet. The results of our proposed method are shown in Table~\ref{tab:res1}. Our method outperforms all the state-of-the-art methods, which achieves 76.77\% in IDF1. Table~\ref{tab:res5} shows the ablation study of different dimension appearance features. The experimental results show that 2048 feature dimensions can achieve the best performance, while a smaller size of dimension decreases the performance. Fig.~\ref{fig:qual_results} shows some of qualitative results, which proves our method can be generalized for different cameras and vehicles. Moreover, the computational complexity of our method is shown in Table~\ref{tab:res4}.

\subsection{Ablation Study}
The ablation study of our method is shown in Table~\ref{tab:res2}. 
There are three components in our framework: SCT, vehicle ReID and TCLM. In terms of SCT, our SCT consists of TNT and TSCT. For vehicle ReID, TA denotes temporal attention and META denotes the use of metadata feature for ReID. The experimental results illustrate how the proposed components enhancing robustness. In Table~\ref{tab:res2}, when replacing TNT with the TSCT, IDF1 based on the TA feature increases by 4.3\% on MTMCT. One significant improvement is to alter the average pooling by TA. Thus, the TSCT and TA are necessary components in our system based on the experimental results. Then, another significant improvement shown in the ablation studies is TCLM. Because not only the transition time constraint between cameras is considered for data association in TCLM but also the spatial information is taken into account at the same time. Since the appearance variance is small in the vehicle ReID, the TCLM can provide significant improvement. Finally, metadata information can further improve the IDF1 by almost 2\%.

\subsection{SCT Performance}

Table~\ref{tab:res3} shows the performance of our approach comparing with those of state-of-the-art SCT methods \cite{wojke2017simple,tang2018single,tang2019moana} in MTMCT. DeepSORT \cite{wojke2017simple} is a Kalman-filter-based online tracking method. Tracklet Clustering (TC) \cite{tang2018single} is an offline method, which is the winner of the AI City Challenge Workshop at CVPR 2018 \cite{naphade20182018}. MOANA \cite{tang2019moana} is the state-of-the-art approach on the MOT Challenge 2015 3D benchmark. In CityFlow dataset, there are three available public detection results (i.e., SSD512 \cite{liu2016ssd}, YOLOv3 \cite{redmon2018yolov3} and Faster R-CNN \cite{ren2015faster}). Since the SSD512 \cite{liu2016ssd} was reported to have the best performance \cite{tang2019cityflow}, therefore we also use SSD512 as our detector directly in our experiments. In our experiments, the adopted metrics of SCT are IDF1 score (IDF1), Multiple Object Tracking Accuracy (MOTA), Multiple Object Tracking Precision (MOTP), Recall and the mostly tracked targets (MT), which are the standard metric for SCT. Based on our experimental results, the proposed TSCT achieves the best performance.

\begin{table}[t]
\centering
\caption{The MTMCT performance for different combinations of the proposed method.}
\vspace{0.5em}
\begin{tabular}{c c c c c | c c c}
\hline 
TNT & TSCT & TA & META & TCLM & IDF1 & IDP & IDR\\
\hline 
\checkmark & & & & & 0.1583 & 0.4418& 0.0959\\
\checkmark & & \checkmark & & & 0.5237 & 0.6816& 0.4221\\
\checkmark & & & & \checkmark & 0.5776& 0.5918 & 0.5779\\
\checkmark & & \checkmark & & \checkmark & 0.7059& 0.6912 & \textbf{0.7211}\\
\checkmark & \checkmark & \checkmark & & \checkmark & 0.7493& 0.8071 & 0.6918 \\
\checkmark & \checkmark & \checkmark & \checkmark & \checkmark & \textbf{0.7677}& \textbf{0.8276} & 0.7159 \\
\hline 
\label{tab:res2}
\end{tabular}
\end{table}

\begin{table}[t]
\centering
\caption{SCT results on CityFlow. }
\vspace{0.5em}
\begin{tabular}{l|c c c c c}
\hline 
Methods & IDF1& MOTA & MOTP & Recall & MT \\
\hline 
DeepSORT \cite{wojke2017simple} & 79.5\% & 68.9\% & 65.5\% & 69.2\% & 756\\
TC \cite{tang2018single} & 79.7\% & 70.3\% & 65.6\% & 70.4\% & 895\\
MOANA \cite{tang2019moana} & 72.8\% & 67.0\% & 65.9\% & 68.0\% & 980\\
\hline 
\textbf{TSCT} & \textbf{90.7\%} & \textbf{83.4\%} & \textbf{75.2\%} & \textbf{87.1\%} & \textbf{1698} \\
\hline 
\end{tabular}
\label{tab:res3}
\end{table}

\section{Conclusion}\label{sec:conclusion}
In this paper, we propose a novel framework for Multi-Target Multi-Camera tracking (MTMCT), which includes traffic-aware single-camera tracking (TSCT), Metadata-Aided Re-Identification (MA-ReID) and also trajectory-based camera link model (TCLM). From our experiments, the proposed method is efficient, effective and robust, which achieves the state-of-the-art performance IDF1 76.77\% in CityFlow dataset for city-scale MTMCT of vehicles.

\section*{Acknowledgment}
This work was partially supported by Electronics and Telecommunications Research Institute (ETRI) grant funded by the Korean government. [20ZD1100, Development of ICT Convergence Technology for Daegu-GyeongBuk Regional Industry]

\ifCLASSOPTIONcaptionsoff
\newpage
\fi



\bibliographystyle{IEEEtran}
\bibliography{main}

%

%

\begin{IEEEbiography}[{\includegraphics[width=1in,height=1.25in,clip,keepaspectratio]{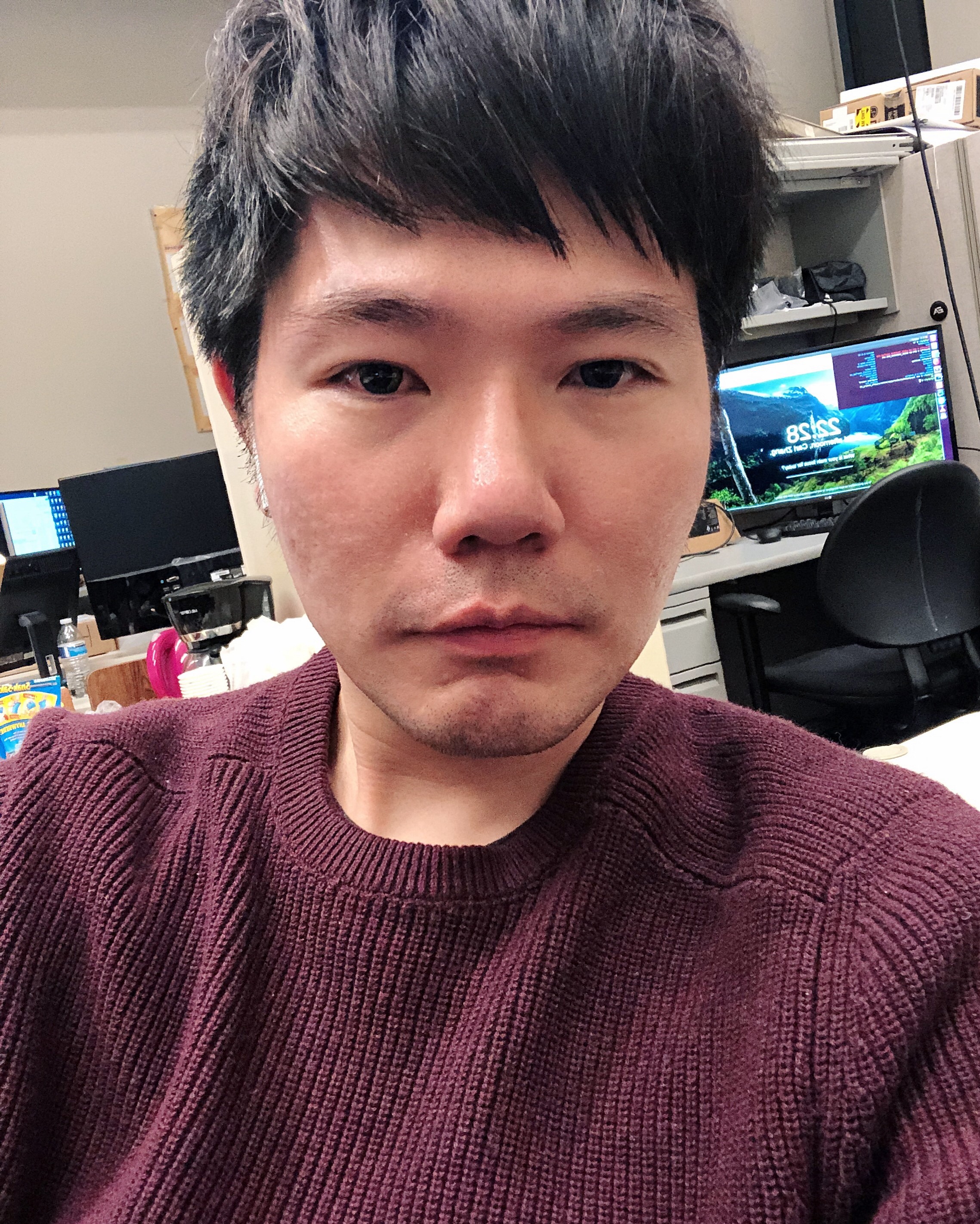}}]{Hung-Min Hsu} received his Ph.D. degree from National Taiwan University in 2017. From 2017 to 2018, He was a postdoctoral fellow in the Research Center for Information Technology Innovation, Academia Sinica. In summer of 2018, Dr. Hsu joined the Department of Electrical and Computer Engineering (ECE) of the University of Washington as a postdoctoral researcher. He has won three CVPR AI City Challenge awards in 2019 and two CVPR Multi-Object Tracking and Segmentation Challenge awards in 2020. His research expertise includes natural language processing, recommendation system and computer vision.
\end{IEEEbiography}

\begin{IEEEbiography}[{\includegraphics[width=1in,height=1.25in,clip,keepaspectratio]{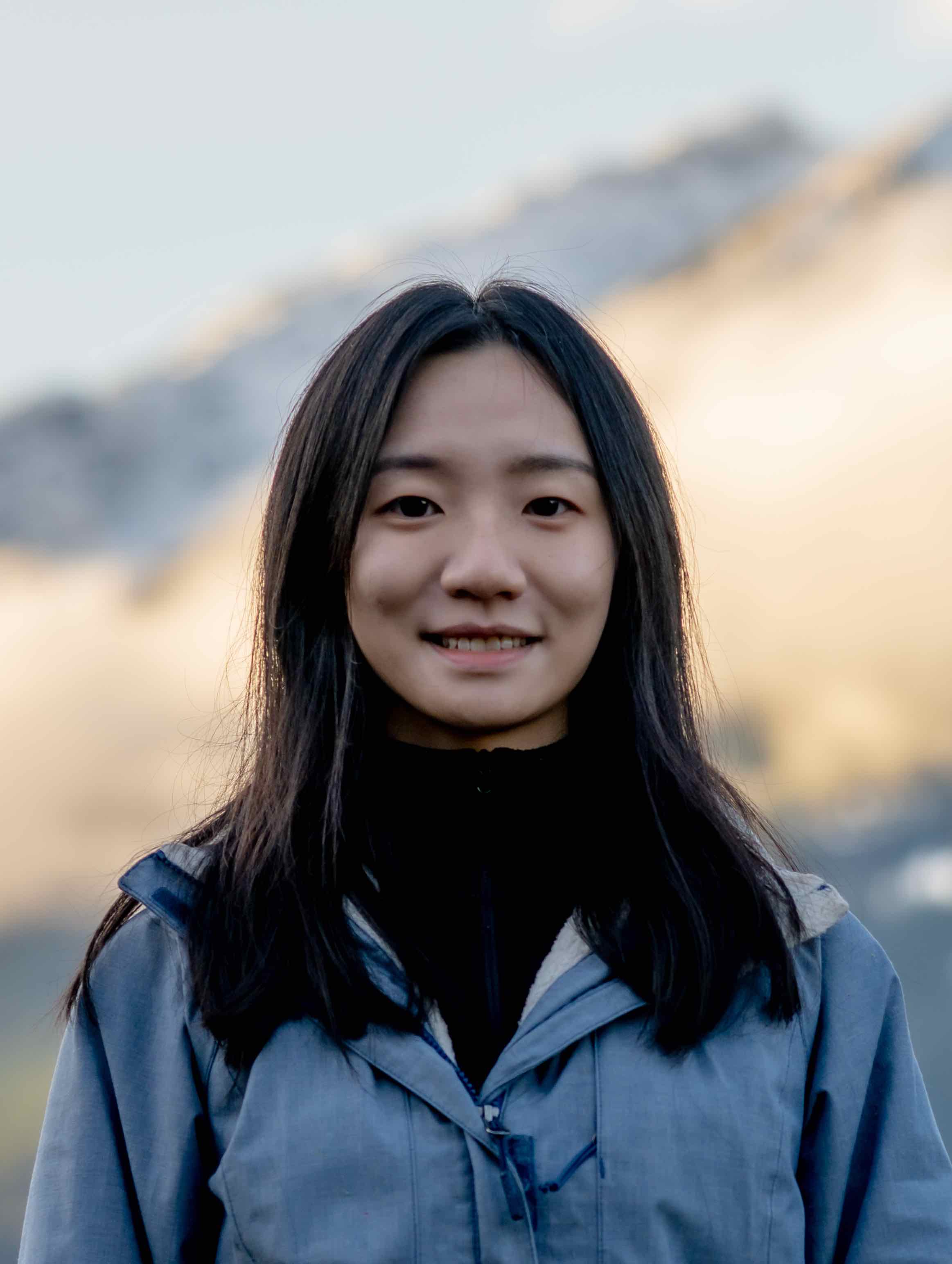}}]{Jiarui Cai}
is a Ph.D. student in the Department of Electrical and Computer Engineering at the University of Washington, Seattle. She received her B.S. degree in Electrical and Information Engineering from Beijing University of Posts and Telecommunications in 2017. Her research includes computer vision and deep learning.
\end{IEEEbiography}

\begin{IEEEbiography}[{\includegraphics[width=1in,height=1.25in,clip,keepaspectratio]{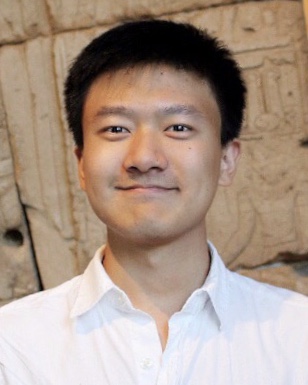}}]{Yizhou Wang}
is a Ph.D. candidate in Electrical and Computer Engineering at the University of Washington, advised by Prof. Jenq-Neng Hwang. He received his M.S. degree in Electrical Engineering in 2018 from Columbia University, advised by Prof. Shih-Fu Chang. He received the B.Eng. degree in Automation at Northwestern Polytechnical University in 2016. His research interests include autonomous driving, computer vision, deep learning, and cross-modal learning. 
\end{IEEEbiography}

\begin{IEEEbiography}[{\includegraphics[width=1in,height=1.5in,clip,keepaspectratio]{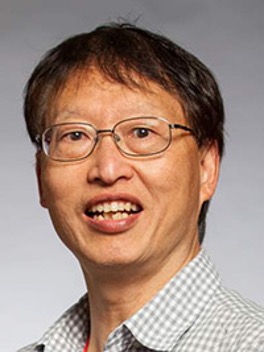}}]{Jenq-Neng Hwang}
received the BS and MS degrees, both in electrical engineering from the National Taiwan University, Taipei, Taiwan, in 1981 and 1983 separately. He then received his Ph.D. degree from the University of Southern California. In the summer of 1989, Dr. Hwang joined the Department of Electrical and Computer Engineering (ECE) of the University of Washington in Seattle, where he has been promoted to Full Professor since 1999. He served as the Associate Chair for Research from 2003 to 2005, and from 2011-2015. He also served as the Associate Chair for Global Affairs from 2015-2020. He is the founder and co-director of the Information Processing Lab, which has won CVPR AI City Challenges awards in the past years. He has written more than 380 journal, conference papers and book chapters in the areas of machine learning, multimedia signal processing, and multimedia system integration and networking, including an authored textbook on ``Multimedia Networking: from Theory to Practice'', published by Cambridge University Press. Dr. Hwang has close working relationship with the industry on multimedia signal processing and multimedia networking. 
Dr. Hwang received the 1995 IEEE Signal Processing Society's Best Journal Paper Award. He is a founding member of Multimedia Signal Processing Technical Committee of IEEE Signal Processing Society and was the Society's representative to IEEE Neural Network Council from 1996 to 2000. He is currently a member of Multimedia Technical Committee (MMTC) of IEEE Communication Society and also a member of Multimedia Signal Processing Technical Committee (MMSP TC) of IEEE Signal Processing Society. He served as associate editors for IEEE T-SP, T-NN and T-CSVT, T-IP and Signal Processing Magazine (SPM). He is currently on the editorial board of ZTE Communications, ETRI, IJDMB and JSPS journals. He served as the Program Co-Chair of IEEE ICME 2016 and was the Program Co-Chairs of ICASSP 1998 and ISCAS 2009. Dr. Hwang is a fellow of IEEE since 2001.
\end{IEEEbiography}

\begin{IEEEbiography}[{\includegraphics[width=1in,height=1.25in,clip,keepaspectratio]{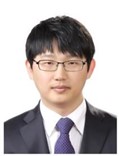}}]{KWANG-JU KIM} received his B.S. degree in electronics engineering from Kyungpook National University (KNU), Daegu, S. Korea in 2010 and M.S. degree in Electrical Engineering from Pohang Institute of Science and Technology (Postech), Pohang, S. Korea, in 2013 and Ph. D. degree in electronics engineering from KNU, Daegu, S. Korea in 2020. From 2013 to 2015, he was a researcher in General Electric Ultrasound Korea (GEUK). Since 2015, he has been with the Electronics and Telecommunications Research Institute (ETRI). His major research interests include computer vision, pattern recognition, and video surveillance.
\end{IEEEbiography}

\end{document}